\documentclass[sigconf]{acmart}

\usepackage{epsfig}
\usepackage{graphicx}

\usepackage{adjustbox}

\usepackage{physics}
\usepackage{tikz}
\usepackage{mathdots}
\usepackage{cancel}
\usepackage{color}
\usepackage{siunitx}
\usepackage{array}
\usepackage{multirow}
\usepackage{tabularx}
\usepackage{extarrows}
\usepackage{booktabs}
\usetikzlibrary{fadings}
\usetikzlibrary{patterns}
\usetikzlibrary{shadows.blur}
\usetikzlibrary{shapes}

\usepackage[utf8]{inputenc}
\usepackage{pgfplots}
\DeclareUnicodeCharacter{2212}{−}
\usepgfplotslibrary{groupplots,dateplot}
\usetikzlibrary{patterns,shapes.arrows}
\pgfplotsset{compat=newest}

\usepackage[]{todonotes}
\usepackage{tabularx}
\usepackage[ruled,vlined,noend]{algorithm2e}

\SetCommentSty{mycommfont}

\usepackage{subcaption}

\copyrightyear{2021} 
\acmYear{2021} 
\setcopyright{acmlicensed}\acmConference[AISec '21]{Proceedings of the 14th ACM Workshop on Artificial Intelligence and Security}{November 15, 2021}{Virtual Event, Republic of Korea}
\acmBooktitle{Proceedings of the 14th ACM Workshop on Artificial Intelligence and Security (AISec '21), November 15, 2021, Virtual Event, Republic of Korea}
\acmPrice{15.00}
\acmDOI{10.1145/3474369.3486862}
\acmISBN{978-1-4503-8657-9/21/11}

\begin{document}

\settopmatter{printacmref=true}

\everypar{\looseness=-1}
\linepenalty=200

\title{Adversarial Transfer Attacks With Unknown Data and Class Overlap}

\author{Luke E. Richards}
\email{richards_luke@bah.com}
\affiliation{%
  \institution{Booz Allen Hamilton}
  \institution{Univ. of MD, Baltimore County}
  \country{USA}
}

\author{Andr\'e Nguyen}
\email{nguyen_andre@bah.com}
\affiliation{%
  \institution{Booz Allen Hamilton}
  \institution{Univ. of MD, Baltimore County}
  \country{USA}
}

\author{Ryan Capps}
\email{capps_ryan@bah.com}
\affiliation{%
  \institution{Booz Allen Hamilton}
  \country{USA}
}

\author{Steven Forsythe}
\email{sforsyth@nvidia.com}
\affiliation{%
  \institution{NVIDIA}
  \country{USA}
}

\author{Cynthia Matuszek}
\email{cmat@umbc.edu}
\affiliation{%
  \institution{Univ. of MD, Baltimore County}
  \country{USA}
}

\author{Edward Raff}
\email{raff_edward@bah.com}
\affiliation{%
  \institution{Booz Allen Hamilton}
  \institution{Univ. of MD, Baltimore County}
  \country{USA}
}

\begin{abstract}
The ability to transfer adversarial attacks from one model (the surrogate) to another model (the victim) has been an issue of concern within the machine learning (ML) community. The ability to successfully evade unseen models represents an uncomfortable level of ease toward implementing attacks. In this work we note that as studied, current transfer attack research has an unrealistic advantage for the attacker: the attacker has the exact same training data as the victim. We present the first study of transferring adversarial attacks focusing on the data available to attacker and victim under imperfect settings without querying the victim, where there is some variable level of overlap in the exact data used or in the classes learned by each model. This threat model is relevant to applications in medicine, malware, and others. 
Under this new threat model attack success rate is not correlated with data or class overlap in the way one would expect, and varies with dataset. This makes it difficult for attacker and defender to reason about each other and contributes to the broader study of model robustness and security. We remedy this by developing a masked version of Projected Gradient Descent that simulates class disparity, which enables the attacker to reliably estimate a lower-bound on their attack's success. 
\end{abstract}

\maketitle

\section{Introduction}

Given an input $\boldsymbol{x} \in \mathbb{R}^d$ and an associated label $y \in \mathcal{Y}$, we wish to learn a model $V(\cdot)$  such that $V(\boldsymbol{x}) = y$. This is achieved using gradient descent on a loss function $\ell(\cdot, \cdot)$, and trained over a dataset $\mathcal{D}^n$ of $n$ samples to minimize the function $\sum_{\forall \boldsymbol{x},y \in \mathcal{D}} \ell(V(\boldsymbol{x}), y)$. 

In this work, we are concerned with untargeted adversarial attacks against the model $V(\cdot)$. For \textit{evasion attacks} specifically, an adversary $A$ will alter $\boldsymbol{x}$ to create a new perturbed sample $\tilde{\boldsymbol{x}} = A(\boldsymbol{x})$. This perturbed $\tilde{\boldsymbol{x}}$ is similar enough to $\boldsymbol{x}$ such that they \textit{should} have the same label, but instead the victim $V(\cdot)$ is fooled so that $V(\tilde{\boldsymbol{x}}) \neq y$. These attacks are surprisingly easy to accomplish. Just as our model $V(\cdot)$ was trained by taking the gradient of the loss with respect to model parameters $\nabla_{\theta} \ell(V(\boldsymbol{x}),y)$, the adversary can create $\tilde{\boldsymbol{x}}$ by taking the gradient of the loss with respect to the input and attempting to maximize the loss.

In the ideal circumstance, we would be able to defend $V(\cdot)$ in the white box scenario, in which the adversary has perfect knowledge of the model $V(\cdot)$, including its parameters, and the training data $\mathcal{D}$. Another common scenario is the gray-box attack, where the adversary $A$ has imperfect knowledge, and wishes to perform a \textit{transfer} attack.

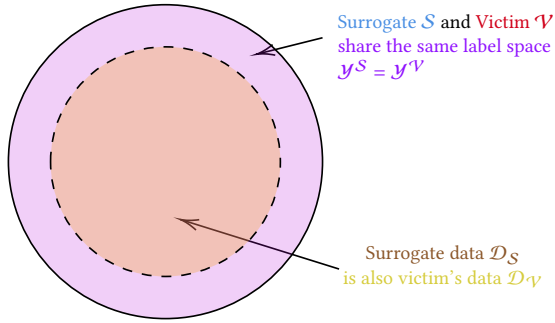
\begin{figure}[!th]
\centering

\adjustbox{max width=0.9\columnwidth}{%
\tikzset{every picture/.style={line width=0.75pt}} %

\begin{tikzpicture}[x=0.75pt,y=0.75pt,yscale=-1,xscale=1]

\draw    (200,30) -- (161.94,39.51) ;
\draw [shift={(160,40)}, rotate = 345.96000000000004] [color={rgb, 255:red, 0; green, 0; blue, 0 }  ][line width=0.75]    (10.93,-3.29) .. controls (6.95,-1.4) and (3.31,-0.3) .. (0,0) .. controls (3.31,0.3) and (6.95,1.4) .. (10.93,3.29)   ;
\draw [line width=0.75]    (210,170) -- (115.91,140.6) ;
\draw [shift={(114,140)}, rotate = 377.35] [color={rgb, 255:red, 0; green, 0; blue, 0 }  ][line width=0.75]    (10.93,-3.29) .. controls (6.95,-1.4) and (3.31,-0.3) .. (0,0) .. controls (3.31,0.3) and (6.95,1.4) .. (10.93,3.29)   ;
\draw  [fill={rgb, 255:red, 189; green, 16; blue, 224 }  ,fill opacity=0.2 ] (10,105) .. controls (10,52.53) and (52.53,10) .. (105,10) .. controls (157.47,10) and (200,52.53) .. (200,105) .. controls (200,157.47) and (157.47,200) .. (105,200) .. controls (52.53,200) and (10,157.47) .. (10,105) -- cycle ;
\draw  [fill={rgb, 255:red, 245; green, 166; blue, 35 }  ,fill opacity=0.3 ][dash pattern={on 4.5pt off 4.5pt}] (35.5,105) .. controls (35.5,66.62) and (66.62,35.5) .. (105,35.5) .. controls (143.38,35.5) and (174.5,66.62) .. (174.5,105) .. controls (174.5,143.38) and (143.38,174.5) .. (105,174.5) .. controls (66.62,174.5) and (35.5,143.38) .. (35.5,105) -- cycle ;

\draw (201,12.4) node [anchor=north west][inner sep=0.75pt]  [color={rgb, 255:red, 144; green, 19; blue, 254 }  ,opacity=1 ]  {$ \begin{array}{l}
\textcolor[rgb]{0.29,0.56,0.89}{\text{Surrogate }}\mathcal{\textcolor[rgb]{0.29,0.56,0.89}{S}}\textcolor[rgb]{0.29,0.56,0.89}{\ }\textcolor[rgb]{0,0,0}{\text{and}}\textcolor[rgb]{0.29,0.56,0.89}{\ }\textcolor[rgb]{0.82,0.01,0.11}{\text{Victim }}\mathcal{\textcolor[rgb]{0.82,0.01,0.11}{V}} \ \\
\text{share the same label space} \ \\
\mathcal{Y^{S} =Y^{V}}
\end{array}$};
\draw (211,152.4) node [anchor=north west][inner sep=0.75pt]  [color={rgb, 255:red, 139; green, 87; blue, 42 }  ,opacity=1 ]  {$\begin{matrix}
\text{Surrogate data} \ \mathcal{D_{S}}\\
\textcolor[rgb]{0.84,0.8,0.24}{\text{is also victim's data}}\textcolor[rgb]{0.84,0.8,0.24}{\ }\mathcal{\textcolor[rgb]{0.84,0.8,0.24}{D}\textcolor[rgb]{0.84,0.8,0.24}{_{V}}}
\end{matrix}$};

\end{tikzpicture}
}
\caption{
Solid circles show the label space, which in purple shows that the victim and surrogate use the exact same label space (perfect intersection). Dashed circles show the data used in training, which again is a perfect union (orange). This is the standard world view explored in prior works. 
}
\label{fig:naive_overlap}
\end{figure}

In the transfer attack scenario the adversary $A$ does not know what model the victim $V(\cdot)$ is using or its parameters, and so they construct a surrogate model $S(\cdot)$. This surrogate can have a different architecture, but is trained in the same manner: $\sum_{\forall \boldsymbol{x},y \in \mathcal{D}} \ell(S(\boldsymbol{x}), y)$. This gives the adversary their own model $S(\cdot)$, which they attack to produce an adversarial sample $\tilde{\boldsymbol{x}}$. The hope, which has held in practice, is that attacks against the surrogate $S(\cdot)$ generalize to the victim $V(\cdot)$ and successful fool it so that $V(\tilde{\boldsymbol{x}})\neq y$. This is how most works have proceeded in the study of this problem, which allows both the surrogate and victim to share the knowledge of the exact same label space (i.e., task) and data, as outlined in \autoref{fig:naive_overlap}. 

The contribution of our work is the recognition of an inconsistency in this threat model: in what realistic circumstance does the adversary $A$ know \textit{nothing} about the victim's model $V$, yet has access to the  same set of class labels  $\mathcal{Y}$ and in non-query scenarios even the \textit{exact} training dataset $\mathcal{D}$? We believe this to be an unrealistic threat model for many real-world scenarios, yet it is the predominant version of the transferable attack scenario studied \cite{Demontis2018,Salman2020,Tramer2017,Papernot2016,236234}. 
As companies develop and label their own corpora, we should expect that the adversary is likely to know \textit{less} about the data than they would the model type.\footnote{Many companies will publish and discuss the types of models they use and develop, but they rarely share their labeled data.} Not all ML models will be exposed to external entities, and may be used entirely for internal purposes within a company. This shared conditioning of the surrogate and victim being trained on the exact same data makes for an unrealistic evaluation. 

\begin{figure}[!th]
\centering

\adjustbox{max width=0.9\columnwidth}{%
\tikzset{every picture/.style={line width=0.75pt}} %

\begin{tikzpicture}[x=0.75pt,y=0.75pt,yscale=-1,xscale=1]

\draw    (330,50) -- (301.41,78.59) ;
\draw [shift={(300,80)}, rotate = 315] [color={rgb, 255:red, 0; green, 0; blue, 0 }  ][line width=0.75]    (10.93,-3.29) .. controls (6.95,-1.4) and (3.31,-0.3) .. (0,0) .. controls (3.31,0.3) and (6.95,1.4) .. (10.93,3.29)   ;
\draw    (290,310) -- (290,0) ;
\draw  [fill={rgb, 255:red, 144; green, 19; blue, 254 }  ,fill opacity=0.2 ] (220,165) .. controls (220,121.19) and (249.66,84.3) .. (290,73.32) .. controls (330.34,84.3) and (360,121.19) .. (360,165) .. controls (360,208.81) and (330.34,245.7) .. (290,256.68) .. controls (249.66,245.7) and (220,208.81) .. (220,165) -- cycle ;
\draw  [fill={rgb, 255:red, 208; green, 2; blue, 27 }  ,fill opacity=0.2 ] (315,70) .. controls (367.47,70) and (410,112.53) .. (410,165) .. controls (410,217.47) and (367.47,260) .. (315,260) .. controls (306.35,260) and (297.97,258.84) .. (290,256.68) .. controls (330.34,245.7) and (360,208.81) .. (360,165) .. controls (360,121.19) and (330.34,84.3) .. (290,73.32) .. controls (297.97,71.16) and (306.35,70) .. (315,70) -- cycle ;
\draw  [fill={rgb, 255:red, 74; green, 144; blue, 226 }  ,fill opacity=0.2 ] (265,70) .. controls (212.53,70) and (170,112.53) .. (170,165) .. controls (170,217.47) and (212.53,260) .. (265,260) .. controls (273.65,260) and (282.03,258.84) .. (290,256.68) .. controls (249.66,245.7) and (220,208.81) .. (220,165) .. controls (220,121.19) and (249.66,84.3) .. (290,73.32) .. controls (282.03,71.16) and (273.65,70) .. (265,70) -- cycle ;
\draw  [fill={rgb, 255:red, 245; green, 166; blue, 35 }  ,fill opacity=0.5 ][dash pattern={on 4.5pt off 4.5pt}] (244.24,169.22) .. controls (244.24,139.74) and (262.32,114.56) .. (287.79,104.58) .. controls (315.56,113.25) and (335.76,139.61) .. (335.76,170.78) .. controls (335.76,200.26) and (317.68,225.44) .. (292.21,235.42) .. controls (264.44,226.75) and (244.24,200.39) .. (244.24,169.22) -- cycle ;
\draw  [fill={rgb, 255:red, 248; green, 231; blue, 28 }  ,fill opacity=0.5 ][dash pattern={on 4.5pt off 4.5pt}] (312.12,100) .. controls (349.61,100) and (380,130.99) .. (380,169.22) .. controls (380,207.45) and (349.61,238.44) .. (312.12,238.44) .. controls (305.19,238.44) and (298.51,237.39) .. (292.21,235.42) .. controls (317.68,225.44) and (335.76,200.26) .. (335.76,170.78) .. controls (335.76,139.61) and (315.56,113.25) .. (287.79,104.58) .. controls (295.34,101.62) and (303.54,100) .. (312.12,100) -- cycle ;
\draw  [fill={rgb, 255:red, 139; green, 87; blue, 42 }  ,fill opacity=0.5 ][dash pattern={on 4.5pt off 4.5pt}] (267.88,240) .. controls (230.39,240) and (200,209.01) .. (200,170.78) .. controls (200,132.55) and (230.39,101.56) .. (267.88,101.56) .. controls (274.81,101.56) and (281.49,102.61) .. (287.79,104.58) .. controls (262.32,114.56) and (244.24,139.74) .. (244.24,169.22) .. controls (244.24,200.39) and (264.44,226.75) .. (292.21,235.42) .. controls (284.66,238.38) and (276.46,240) .. (267.88,240) -- cycle ;
\draw [line width=0.75]    (320,310) -- (300.39,211.96) ;
\draw [shift={(300,210)}, rotate = 438.69] [color={rgb, 255:red, 0; green, 0; blue, 0 }  ][line width=0.75]    (10.93,-3.29) .. controls (6.95,-1.4) and (3.31,-0.3) .. (0,0) .. controls (3.31,0.3) and (6.95,1.4) .. (10.93,3.29)   ;
\draw [line width=0.75]    (190,270) -- (219.11,211.79) ;
\draw [shift={(220,210)}, rotate = 476.57] [color={rgb, 255:red, 0; green, 0; blue, 0 }  ][line width=0.75]    (10.93,-3.29) .. controls (6.95,-1.4) and (3.31,-0.3) .. (0,0) .. controls (3.31,0.3) and (6.95,1.4) .. (10.93,3.29)   ;
\draw    (420,250) -- (351.23,161.58) ;
\draw [shift={(350,160)}, rotate = 412.13] [color={rgb, 255:red, 0; green, 0; blue, 0 }  ][line width=0.75]    (10.93,-3.29) .. controls (6.95,-1.4) and (3.31,-0.3) .. (0,0) .. controls (3.31,0.3) and (6.95,1.4) .. (10.93,3.29)   ;

\draw (303,28.4) node [anchor=north west][inner sep=0.75pt]  [color={rgb, 255:red, 144; green, 19; blue, 254 }  ,opacity=1 ]  {$\text{Feasible attack space} \ \mathcal{Y^{S} \cap Y^{V}}$};
\draw (231,13.5) node  [font=\large,color={rgb, 255:red, 74; green, 144; blue, 226 }  ,opacity=1 ]  {$\text{Surrogate }\mathcal{S}$};
\draw (332.5,13.5) node  [font=\large,color={rgb, 255:red, 208; green, 2; blue, 27 }  ,opacity=1 ]  {$\text{Victim }\mathcal{V}$};
\draw (293,312.4) node [anchor=north west][inner sep=0.75pt]  [color={rgb, 255:red, 245; green, 166; blue, 35 }  ,opacity=1 ]  {$\mathcal{D_{S} \ \cap D_{V}}$};
\draw (412,168.4) node [anchor=north west][inner sep=0.75pt]  [color={rgb, 255:red, 144; green, 19; blue, 254 }  ,opacity=1 ]  {$\mathcal{\textcolor[rgb]{0.82,0.01,0.11}{Y^{V}}}$};
\draw (143,162.4) node [anchor=north west][inner sep=0.75pt]  [color={rgb, 255:red, 74; green, 144; blue, 226 }  ,opacity=1 ]  {$\mathcal{\textcolor[rgb]{0.29,0.56,0.89}{Y^{S}}}$};
\draw (111,275.4) node [anchor=north west][inner sep=0.75pt]  [color={rgb, 255:red, 139; green, 87; blue, 42 }  ,opacity=1 ]  {$\begin{matrix}
\text{Surrogate data} \ \mathcal{D_{S}}\\
\text{collected by adversary}
\end{matrix}$};
\draw (427,276.5) node  [color={rgb, 255:red, 215; green, 203; blue, 61 }  ,opacity=1 ]  {$\begin{matrix}
\text{Victim's training}\\
\text{ Data} \ \mathcal{D_{V}}
\end{matrix}$};

\end{tikzpicture}
}
\caption{In a transfer attack where the adversary attacks their own surrogate model $S(\cdot)$ in order to evade a victim's model $V(\cdot)$. We study two factors that impact the attack success. The overlap between the data $\mathcal{D_S}$ used to train $S(\cdot)$ and $\mathcal{D_V}$ used to train $V(\cdot)$  (dashed lines) exists in the sub-space of the overlap in label output spaces (solid lines) $S(\cdot): \to \mathcal{Y^S}$ and $V(\cdot): \to \mathcal{Y^V}$. These two factors may vary in their degree of overlap, and data may have no overlap. 
}
\label{fig:overlap}
\end{figure}
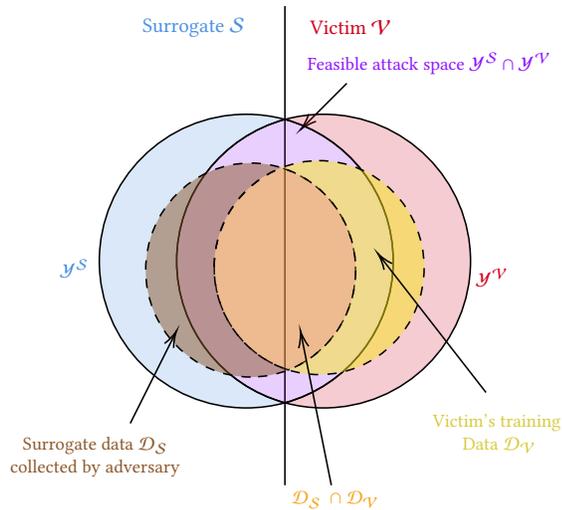

Instead, we study a new version of the transfer attack scenario summarized in \autoref{fig:overlap}. For simplicity we will allow the surrogate $S$ and victim $V$ to use the same architecture when using different datasets $\mathcal{D_S}$ and $\mathcal{D_V}$ each with $n$ samples. Since it is common practice to still use publicly available data, we will study scenarios where ranging from no overlap ($\mathcal{D_S} \cap \mathcal{D_V} = \emptyset$), partial overlap ($|\mathcal{D_S} \cap \mathcal{D_V}| < n$), and complete overlap ($\mathcal{D_S} = \mathcal{D_V}$). In addition we note that companies may not fully expose all of the classes they might detect, so we will also study how the degree of overlap between different output spaces  $\mathcal{Y}^S$ and $\mathcal{Y}^V$ in the same fashion. In doing so, we discover that the lack of shared underlying data can significantly reduce the success of transfer attacks from the surrogate $S(\cdot)$ to the victim $V(\cdot)$ in unexpected ways, as it does not correlate with the degree of overlap in the way one may intuitively expect. This makes it difficult for the adversary to determine the likelihood an attack succeeds, and thus to plan said attack. We also show how state-of-the-art adversarial training can actually help improve the attacker's success rate due to a kind of over-fitting. To restore behaviors to an easier to understand and interpret set of patterns, we develop an augmented ``masked surrogate'' attack that the adversary can use which removes the variance caused by imperfect class overlap, making it possible for $A$ to again attack victim $V(\cdot)$ with confidence.  

To help motivate this problem, consider the example of pain in a medical context. A hospital using a machine learning model to try and detect different types and levels of pain from facial expressions can be useful for triage and patient care\cite{Broome2019,Lucey2009,Hammal2012,Ashraf2009,Martinez2017}. It would not be queryable by outsiders to attack, but patients engaging in drug-seeking behavior\footnote{For their own use or illegal re-sale. The later group potentially well funded and motivated.} would desire to attack such a system \cite{Finlayson,dr_ai_long}. An adversary could realistically create their own dataset, but because there are many different medical taxonomies and definitions around pain \cite{Treede2018}, they would not likely know the exact set of labels used by their victim.

The rest of our paper is organized as follows. In \autoref{sec:related} we will discuss related work which has used the unrealistic version of our threat model.
In \autoref{sec:approach} we contribute a rigorous threat model for this unrecognized, and realistic  scenario, and a new masked PGD attack for the threat model. 
In \autoref{sec:experiments} and \autoref{sec:results} we contribute an extensive empirical study of the threat model showing that current adversarial methods and intuition do not hold under uncertain class and data overlap, making attacks unreliable. Our new masked PGD will restore the attacker to an intuitive behavior that allows effective attack against $V(\cdot)$. 
Finally we will conclude in \autoref{sec:conclusion}.

\section{Related Work} \label{sec:related}

While adversarial attacks on machine learning algorithms have received increased attention in the last few years, they have been a research area for over a decade \cite{Biggio2017} and have proven to be challenging to defend against \cite{Carlini:2017:AEE:3128572.3140444}. The threat of adversarial attacks applies broadly to many classification use cases, for example malware detection \cite{Kolosnjaji2018AdversarialExecutables} and facial recognition \cite{Shan2020Fawkes:Models}, as well as to regression tasks \cite{Nguyen2018AdversarialRegularization}.
In this work, we are interested in the transfer of adversarial attacks. There are two larger groups that transfer attacks have fallen into based on whether they use queries against the victim $V(\cdot)$ or not. Our work extends the study of threat models that do not query the victim. 

Many have looked at the transferability of adversarial attacks that do not rely on querying the victim model \cite{Kariyappa2019,8953425,Zhou2018,Salman2020,Tramer2017,8953423,dong2018boosting} and built theory of how to improve transfer success rate \cite{Demontis2018}. All of these cited works assume, or do not discuss, the adversary having perfect knowledge of the victim's classes (i.e., $\mathcal{Y^S} = \mathcal{Y^V}$). 
The few works that have recognized that shared data ($\mathcal{D_S} = \mathcal{D}_V$) is an unrealistic assumption have not studied how the degree of data overlap may impact transfer success,
and instead replace the assumption with the case that the adversary knows how much of their data overlaps. Prior work \cite{dong2018boosting, cheng2019improving, dong2019evading} modified gradient-based attacks to optimize transferability between models without examining data overlap. 
These works used mostly 10-100 attack iterations, and found that more iterations reduced the impact of their results. When using 250 to 1,000 as we do in our work, we found these approaches had no impact on results (e.g., see \autoref{fig:MINIIMGNETSUCESS250_MIFGSM}).

Our work contributes the realization that unknown class and data overlap is a likely scenario, and that if you can't query your victim you may not have certainty about class overlap either. 
Some have intentionally made sure there is no data overlap, but not tested the impact of no or partial overlap \cite{10.1109/SP.2014.20}. 
As far as we are aware, we are the first to study the impact of class overlap and the first to study how the degree of data overlap impacts transfer success, and the interaction of these two factors. 

Many black-box attacks  query the victim \cite{Papernot2016,10.1145/3052973.3053009,Demontis2018} with new data points to get them labeled. This means that each label pair in the surrogate's corpus $\boldsymbol{x}, y \in \mathcal{D}_\mathcal{S}$ is really $\boldsymbol{x}, V(\boldsymbol{x})$. While the attacker is limited based on the number of queries considered, they are implicitly gaining information about the entire training set $\mathcal{D}_\mathcal{V}$ that was used to create $V(\cdot)$. 
The query based black-box attacks are realistic but do not cover all circumstances (e.g., models used internally or which do not provide immediate/direct feedback).

One of the only works we are aware of the recognize the adversary may have varied degrees of knowledge or overlap with $\mathcal{D}_\mathcal{V}$ was by \cite{217486}. Their study however was focused on only poisoning based attacks, and did not explore the broader impact of the degree of overlap $|\mathcal{D_S} \cap \mathcal{D_V}|/\max(|\mathcal{D_S}|,|\mathcal{D_V}|)$. Instead they choose to allow the adversary to know 70\% of the training corpus or 100\%. Instead we will test a spectrum of overlaps of training data and class labels.

To the best of our knowledge, no work has yet considered that the adversary may not know the exact set of classes $\mathcal{Y^V}$ used by the victim, and so must use an overlapping but different set of labels $\mathcal{Y^S}$ for their surrogate though the scenario is realistic. Determining what family of malware (i.e., the class) a given executable belongs to is important functionality, yet all anti-virus products have wildly different sets of labels they recognize with different definitions, which are constantly changing as new families are detected \cite{Sebastian2016,Seideman2014,Sebastian2020}. The malware example is particularly motivating because it involves a real-live adversary (the malware author) who actively wants to avoid detection. Our model of unknown class overlap is also common to industry practice. Many companies will have alpha/beta tests or ``phantom'' deployments of hidden features for testing, which could include new class labels that have not been exposed. Even under normal circumstances, companies do not always disclose all classes they recognize or precisely how many exist.\footnote{e.g., Clarifai's current documentation gives a lower bound and no comprehensive list for the number of classes available for many of their models. This behavior is relatively common amongst commerically available APIs.} Real-world production models can have millions and even 100 million classes in use \cite{10.1145/3394486.3403368,10.1145/3289600.3290979}, making it impractical for an adversary to likely model all available classes. While we do not test to this extreme, we believe this provides further evidence  that assuming a perfect model of the output space $\mathcal{Y^V}$ is not always realistic, and thus worthy of our study.

\section{Approach}\label{sec:approach}

As recommended by \cite{Biggio2014} we first make explicit the threat model that defines our adversary's goals, capabilities, and knowledge. We assume there is a victim model $V(\cdot)$ that they wish to provide altered inputs $\tilde{\boldsymbol{x}} = A(\boldsymbol{x})$ such that $V(\tilde{\boldsymbol{x}})$ will produce incorrect predictions. The adversary is able to send altered inputs $\tilde{\boldsymbol{x}}$ to the victim, but is not able to observe a direct or indirect response. The adversary does not know the training data $\mathcal{D_V}$ used by the victim, or the specific classes-- it is only inferred guesses or availability that they should overlap to an unknown degree  (as we have established in \autoref{sec:related}, this is realistic for cases of malware, medical, and other applications). The adversary has the capability to perturb inputs by up to $\epsilon$ under the $\infty$ norm, such that $\|\tilde{\boldsymbol{x}}-\boldsymbol{x}\|_\infty \leq \epsilon$ by attacking their own surrogate model $S(\cdot)$ trained on an independently built corpus $\mathcal{D_S}$ with an independent label set $\mathcal{Y^S}$. Pessimistically we assume the adversary's surrogate model $S(\cdot)$ uses the exact same architecture as the victim's $V(\cdot)$ (given wide re-use of ``backbone'' networks such as ResNet, this is a reasonable assumption). The adversary needs to meet a minimum attack success rate of $p\%$ for their attack to be viable, and only has one chance to launch their attack campaign\footnote{The assumption being a failed attack alerts the defender, who can thus make changes or react in some unfavorable way, that is ultimately a detriment to the attacker's goals.}.

Our restrictions on the adversary's knowledge of the data and classes may seem extreme. This requires the adversary to invest time into understanding their victim, building their own corpus and expected label set, and then producing perturbed inputs to send to the victim. This scenario is realistic to many situations, and occurs in biometric recognition~\cite{Arslan2017}, medicine \cite{Finlayson,dr_ai_long}, and malware detection \cite{Anderson2018,Egele2017,Raff2020a}
 are all domains where the specific class labels may vary between users but have predictable high-level goals, and data is tightly held (for privacy or competitive advantage), and models are often kept private or not fully exposed to users/consumers. 

Within the threat model, we assume that there are varying levels of intersection between the data points of  $\mathcal{D_S}$ and  $\mathcal{D_V}$. Within this analysis we make splits based on a number of shared classes for image classification $\mathcal{Y} = {y_1, y_2, ... , y_N}$ where each model is trained using classes a subset of $\mathcal{Y}$, the surrogate having $\mathcal{Y}^S$ and the victim $\mathcal{Y}^V$. Where the shared classes of $\mathcal{Y^S} \cap \mathcal{Y^V} = \mathcal{Y^O}$  represents the overlap in shared classes. We observe various sizes of intersection/overlap $|\mathcal{Y^{O}}| $  within the work respective to the size of $|\mathcal{Y}|$. Because we test with varying levels of class overlap, each model ($V(\cdot)$ and $S(\cdot)$) will have half as many output classes. For example, CIFAR-100 tests would only have 50 classes so that we can vary the number of shared classes from 2 to 100\% without changing the number of classes. 

Within that intersection of shared classes, we split the intersection further by dividing the data points of classes $\mathcal{Y^{O}}$  based on a varying number of data partitions such that the data points $\mathcal{D_{Y^{O}}}$ into subsets such that $\mathcal{D_S} \neq \mathcal{D_V}$. Thus the surrogate model is trained on $\mathcal{D_S} = \mathcal{D_{Y^{S}}} \cup \mathcal{D_{Y^{O}}}$ and the victim model is trained on $\mathcal{D_V} = \mathcal{D_{Y^{V}}} \cup \mathcal{D_{Y^{O}}}$. We analyze where there is no overlap in shared data points but shared classes exist $\mathcal{D_S} \cap  \mathcal{D_V} = \emptyset$, the case in which there is partial overlap of data points $|\mathcal{D_S} \cap \mathcal{D_V}| < n$ , and finally the case in which there it total overlap $\mathcal{D_S} = \mathcal{D_V}$.  

To better study the impact of $|\mathcal{D_{Y^O}}| / \max(|\mathcal{D_S}|,|\mathcal{D_V}|)$ the shared data ratio  between the surrogate $S(\cdot)$ and victim $V(\cdot)$, we will not use all of the original training data. Instead, each model will be trained only on half of the original samples (e.g., using 30,000 out of the 60,000 training samples of MNIST). This is done so that we can test a shared class overlap of $\{0, 25, 50, 75, 100\}\%$ while keeping the total training size constant, eliminating it as a factor of variation. For example, 0\% overlap (i.e., $|\mathcal{D_{Y^O}}|/\max(|\mathcal{D_S}|,|\mathcal{D_V}|) = 0$),  would mean all of the original data is used, half by the surrogate $S$ and the other half by the victim $V$. For 100\% overlap only half of the original data is used by both $S(\cdot)$ and $V(\cdot)$. 
This way all models trained have a corpus of the same total size. A consequence of this design choice is our results are not directly comparable with prior studies on the same datasets, since our models are trained on only half as much data. 

Once both models are trained on their respective datasets, we find an adversarial attack $A$ in which the input  $\boldsymbol{x}$  is perturbed to be $\tilde{\boldsymbol{x}} = A(\boldsymbol{x})$ for the surrogate model  $S(\cdot)$. We then transfer this attack $A$ to the victim model $V(\cdot)$, which $A$ is not optimized for. The success of such a transfer attack is measured by the original accuracy of the victim model on the shared held out test data $\mathcal{D_{Y^{O}}}$ subtracted from the new accuracy after $A$ perturbs $\mathcal{D_{Y^{O}}}$, $A(\mathcal{D_{Y^{O}}})$. 

\begin{equation}
\operatorname{success} = \operatorname{accuracy}(V(\cdot), \mathcal{D_{Y^{O}}}) -  \operatorname{accuracy}(V(\cdot), A(\mathcal{D_{Y^{O}}}))
\end{equation}
This metric allows us to examine how dependent the adversarial attack $A$  is on the data of the surrogate $\mathcal{D_S}$.  Common thought from the machine learning community would assume the closer the overlap in data distribution there is between $\mathcal{D_S}$ and $\mathcal{D_V}$ the more successful the attack will be. We examine how well this initial hypothesis holds with empirical evidence in the following section.

\subsection{Impact of Adversarial Training} 

We examine a variation of the threat model in which both the surrogate model and the victim model are hardened to adversarial attacks during training. Each model then is exposed to adversary examples $\tilde{\boldsymbol{x}}_S$ generated by $A_S$ for model $S(\cdot)$ to have $S(\tilde{\boldsymbol{x}}_S) \not= y$ . The same process is used for $V(\cdot)$ to emulate the victim also attempting to safeguard the model. Under normal study this results in improved accuracy under attack, at some variable decrease in accuracy on clean samples. Our results will show this behavior does not hold under our new threat model.

\subsection{Removing Class Uncertainty}

Finally, to explore the dependence of classes being present within the surrogate model $\mathcal{Y^S}$, we introduce a modified PGD attack, 
Algorithm 1, 
that runs $T$ additional iterations with a randomly generated mask ($M$) over the logits of the class output. 
Randomly masking classes makes it such that the adversary cannot rely on any ``easy'' classes (e.g., two different breeds of dog) to obtain successful attacks. This operation is in a manner mechanically similar to Dropout~\cite{Srivastava2014} and Expectation over Transformation (EoT) \cite{Athalye2017}. Unlike Dropout our goal is to make the attacker more robust, and unlike EoT we are not differentiating through a defence of the victim (EoT's original purpose), but instead circumventing the adversary's uncertainty about the victim through simulation. 
We then compute the delta $\delta$ per the PGD attack $T$ times and then average it to have a more generalized delta, $\delta_{avg}$, that will not be dependent on $\mathcal{Y^S}$. We believe this to be a better strategy for attacking the victim model with no guarantee of any of the surrogate model classes $\mathcal{Y^S}$ being present in victim's $\mathcal{Y^V}$.

\begin{algorithm}[!h]
\SetAlgoLined
 \For(\tcp*[h]{Samples to attack}){$n = 1 ... S$}  
 {
    \For(\tcp*[h]{PGD iterations}){$i = 1 ... N$}{
     $\delta_{avg} = 0$
     
        \For(\tcp*[h]{Mask iterations}){$j = 1 ... T$}{
            $M$ = random mask \tcp*{Sample a mask for logits} 
            
            $x_p = f_\theta(x) \cdot M$ \tcp*{Apply mask, M to model output}
            
            $\delta = \delta + \alpha \cdot \operatorname{sign}(\nabla_\delta \ell(x_p + \delta, y))$
            
            $\delta_{avg} = \delta_{avg} + \max(\min(\delta, \epsilon), -\epsilon)$
        }  
    }
     $\delta_{avg} = \delta_{avg} / T$ 
  }
 \caption{Masked PGD Attack: N number of iterations for PGD, S number of examples in dataset, T number of mask iterations}
 \label{algo:maskedPGD}
\end{algorithm}

\subsection{Computational Considerations}
\label{compcost}

A single model on CIFAR-100 takes on average 1.25 hours to train on a single graphics processing unit (GPU). For each experiment we must train the surrogate model and the victim model from scratch to preserve control for data overlap.  Each with a change in the number of classes shared and data, for CIFAR-100 looking at 5 number of classes each at 5 different percentages of data shared creates 25 trains of the surrogate model and then 25 trains for the victim model. This totals in 50 total trained models. This totals to 62.5 hours for each training experiment. Due to the random selection of the shared classes and shared data, we run each experiment 5 times to account for outliers and average the performance. This increases to 312.5 GPU hours in training. This is not including the total time it takes to find and run the attack on the victim network, which in the next section we choose to take an extremely high number of iterations to properly capture the potential of the attack for our threat model. The attack cost for our masked experiments take another $\approx$ 3,700 GPU hours due to the iterative masking process. The results presented constitute months of our entire compute capacity. 

Performing the same study, with multiple runs to average the behavior, would take an estimated $\approx40,000$ GPU hours ($\approx 4.5$ years), is simply beyond our resources and financial means. We have results for a reduced and single-run (no averaging) result on ImageNet in \autoref{sec:appendix_imagenet} as a spot check, which supports our overall conclusions derived from the more tractable CIFAR and Mini-ImageNet datasets.

\section{Experiments} \label{sec:experiments}

We examine the success rate of transferring an attack to a victim model $V(\cdot)$ from a surrogate model $S(\cdot)$ with varying levels of shared tasks/data. We define tasks as shared-classes and data as the percentage of images within that intersection of shared-classes that both models are trained on. 
The training data is split based on the number of shared classes leaving the non-intersecting classes to be split between the two models. From the shared class data, we then split based on a percentage shared. We then train two networks on each of these subsets.

Within our extended experiments, we focus on Mini-ImageNet since this dataset is larger, in terms of both classes and data, and diverse in the type of images and classes. We find this to be a reasonable size for our extensive computational costs of experiments compared to ImageNet (see \autoref{compcost}). Two variations are tested, one in which both surrogate and victim models are adversarially hardened during training and a novel variation of projected gradient descent (PGD) attack in which we mask classes to increase generalization of transfer. 

In summary we conduct the following experiments:
\begin{itemize}

  \item An analysis of success of transfer from surrogate model to victim model at various levels of shared classes and data points.
  \item Analysis of the effects adversarial hardening during training of both models has on the success based on varying levels of shared classes and data points. 
  \item A novel adversarial attack meant to generalize without relying on a single class present in the surrogate model using masking based targets. 
\end{itemize}

\begin{figure*}[!h]
\centering
\includegraphics[width=0.95\textwidth, scale=1]{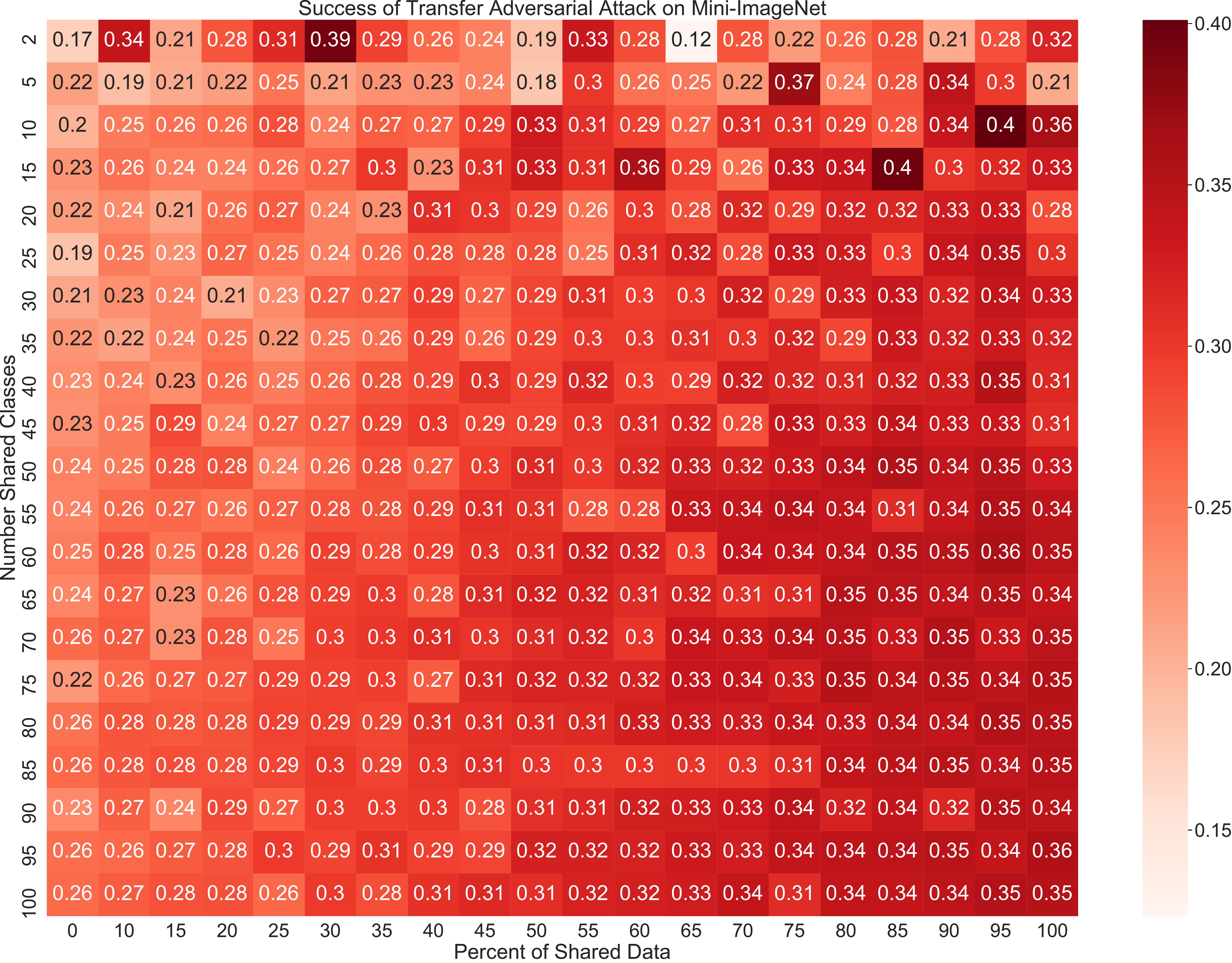}
\caption{This figure shows a fine-grained experiment run for a sub-sample of ImageNet using standard PGD. Examining the attack success at such scale allows greater visualization of the variance we see at a smaller scale and the relation between the two factors. With multiple runs per cell to account for the variance, this experiment took approximately 3,168 GPU hours. This dataset shows high variance when the number of shared classes $\leq 25$.}
\label{fig:LARGEMINIIMGNETSUCESS250}
\end{figure*}

\subsection{Standard Adversarial Transfer Attack}

We use ResNet architectures \cite{he15deepresidual}, 
a ubiquitous backbone architecture,
for all experiments. We choose to match the size of model to the size of dataset being learned. We used ResNet-18 for Fashion-MNIST, ResNet-50 for CIFAR-10, and Wide ResNet-50 for CIFAR-100. Each network is then modified to have a final linear layer that is only the size of the number of classes per model. We used a ResNet-18 model for Mini-ImageNet for computational cost considerations.

The models are trained using Adam optimizer \cite{Kingma2015} with a learning rate of 0.0004 selected on empirical results, with varying batch sizes based on dataset: 12 for Fashion-MNIST, 12 for CIFAR-10, 128 for CIFAR-100, and 128 for mini-ImageNet. The number of epochs for each model were 10 for Fashion-MNIST, 20 for CIFAR-10, and 200 for both CIFAR-100 and mini-ImageNet. For CIFAR-100 and Mini-ImageNet we used a learning rate decay at epochs 60, 120, and 160 with gamma 0.2. 

Once each model is trained on the respected data-subset a perturbation input $\tilde{\boldsymbol{x}} = A(\boldsymbol{x})$ is found using projected gradient descent (PGD) \cite{Kurakin2017} by attacking the surrogate model $S(\cdot)$ with $L_\infty$ normalization and an $\epsilon$ of 0.3 with a PGD step size of 0.01. 
Following best practices we use as many PGD steps as computationally feasible \cite{Athalye2018}, in our case 1,000 steps (though we found 100 sufficient in most cases). 
Then we apply that computed perturbation, $A$ to the victim model, $V(\cdot)$. We measure the success of the transferred attack based on varying combinations of shared-classes and percentage shared. 
Our results will show that having less shared data has a non-trivial impact on attack success rate, and that fewer shared classes has a large impact on the adversary's estimate of attack success rate and could lead to over/under confidence based on a computationally constrained attacker. We showcase this standard threat model with varying levels of attack with a small step size (see \autoref{fig:LARGEMINIIMGNETSUCESS250}). This allows us to broadly view the purpose of this experiment to examine the effects these factors have on transfer attacks. This example demonstrates in greater fidelity the change in transfer attack success rate as the percentage of shared classes and data change and interact in non-linear ways. We note that the specific nature of the pattern changes from one dataset to the next, and we do not have the computational capacity to repeat this high fidelity testing for each dataset. Our Masked PGD attack helps alleviate the uncertainty of what behavior will occur by removing the impact of the number of shared classes, and results in a consistent and smooth behavior that more shared data leads to high attack success rate (which is not true under standard PGD for all datasets, including \autoref{fig:LARGEMINIIMGNETSUCESS250} itself).

\subsection{Adversarial Hardening }

Within the threat model, we also have to account for the victim model creators attempting to mitigate the success of attacks, and attackers trying to create more robust attacks~\cite{Salman2020}. 
We do this by performing adversarial training on both $S(\cdot)$ and $V(\cdot)$, and conducting the same grid of transfer attack testing, to determine how viable transfer attacks are when the adversary has the limited visibility of our specified threat model. 
Due to the number of train/test runs, we use the work of \cite{Wong2020} for computationally efficient hardening of both networks during training. Within this method, the network is trained using a randomly initialized fast gradient sign method (FGSM) attack which was found to be as effective as the PGD attack.
Our prior expectation is that this would hamper the adversary, but our results show the drop in attack success rate is smaller than we might have expected. 

\subsection{Masked Transferred Attacks}

The masked PGD attacks were too expensive to run to 1,000 iterations, so only 250 iterations were done instead. This number appears to still be reasonable, and was tested by running the standard attack for only 250 PGD iterations and no difference in results were observed. We note that in the masked PGD case, each iteration requires 100 sub-iterations (averaging over the random masks $M$) for an effective total of 250,000 gradient calculations. 
Therefore we feel this decrease is a reasonable change.  We will show that this masked attack removes the number of shared classes as a source of variance, allowing the attacker to confidently estimate their attack success rate.

\section{Results} \label{sec:results}

Our results shine light on behaviors outside of the intuition given the factors of shared data and classes.
A priori one would expect that an attack's success should drop when class and data overlap decrease due to a lack of information, and that adversarial training would improve the victim's accuracy under attack \cite{Athalye2018,Madry2018}. Our results show that these intuitions do not hold in practice: there is a dataset dependant pattern of correlation between class/data overlap and attack success rate, and adversarial training of the victim can counter-intuitively improve the attacker's success rate. These make it difficult for both attackers and defenders to reason about their degree of vulnerability. 
This adds a human-factor within the threat model of attack uncertainty minimization.
Our masked PGD attack Algo. \ref{algo:maskedPGD} restores the attacker to a more consistent behavior, removing the variability with respect to class overlap and restoring standard expectations to the attacker's success rate decreasing as the amount of data shared with the victim decreases. Assuming an attacker would not engage in a campaign with high uncertainty, this allows the attacker (and thus, defender) to better quantify and determine their success or risk respectively. For readers viewing, we have used red for success values, blue for standard deviation, and green for differences between two  methods of attack or training. 

\subsection{Standard Adversarial Transfer Attack}

Looking first at the smaller dataset behaviors (CIFAR-10 and FashionMNIST), we see an overall trend of higher success when there is more overlap in both data and classes (as originally expected). Within both however, there was an interesting dip in success with 50\% shared data and 2 classes before rising again. Within CIFAR10, the model actually had more success with 0\% overlap opposed to 100\% overlap. Fashion-MNIST however had more consistency with success increasing as overlap increases.

For CIFAR10 (\autoref{fig:CIFAR10CFNMTX}) the correlation coefficient between number of shared classes and success was 0.6578, for Fashion-MNIST (\autoref{fig:FashionMNISTCFNMTX}) it was 0.8213. These values point to a similar relationship, however the peak within sharing 50\% of the classes in the CIAFAR10 data skewed this relationship.  For the correlation coefficient for the percentage of data shared, CIFAR10 had  -0.4310 and Fashion-MNIST had 0.3238. 
The negative coefficient is counter-intuitive going against the a priori expected trend on the small datasets, but as we look at larger datasets with more classes the behavior will regress further.

\begin{figure}[!h]
\centering
\vspace{1ex}
\includegraphics[width=1\columnwidth]{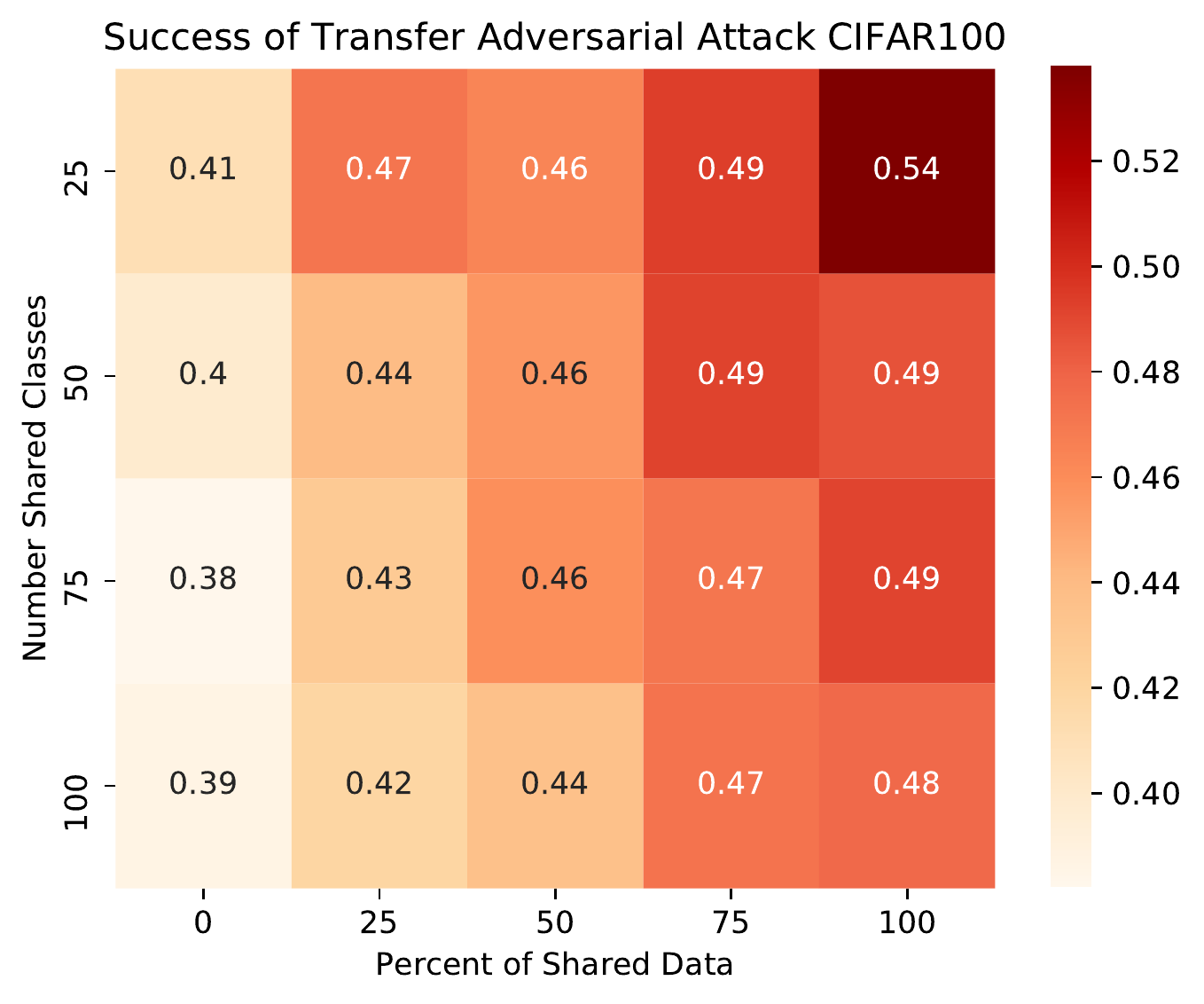}
\vspace{-2ex}
\caption{Success of transfer attack dependent on variations of shared classes and percentage of data shared for CIFAR-100.}
\label{fig:CIFAR100CFNMTX}
\end{figure}

\begin{figure}[!th]
\centering
\vspace{1ex}
\includegraphics[width=1\columnwidth, scale=2.5]{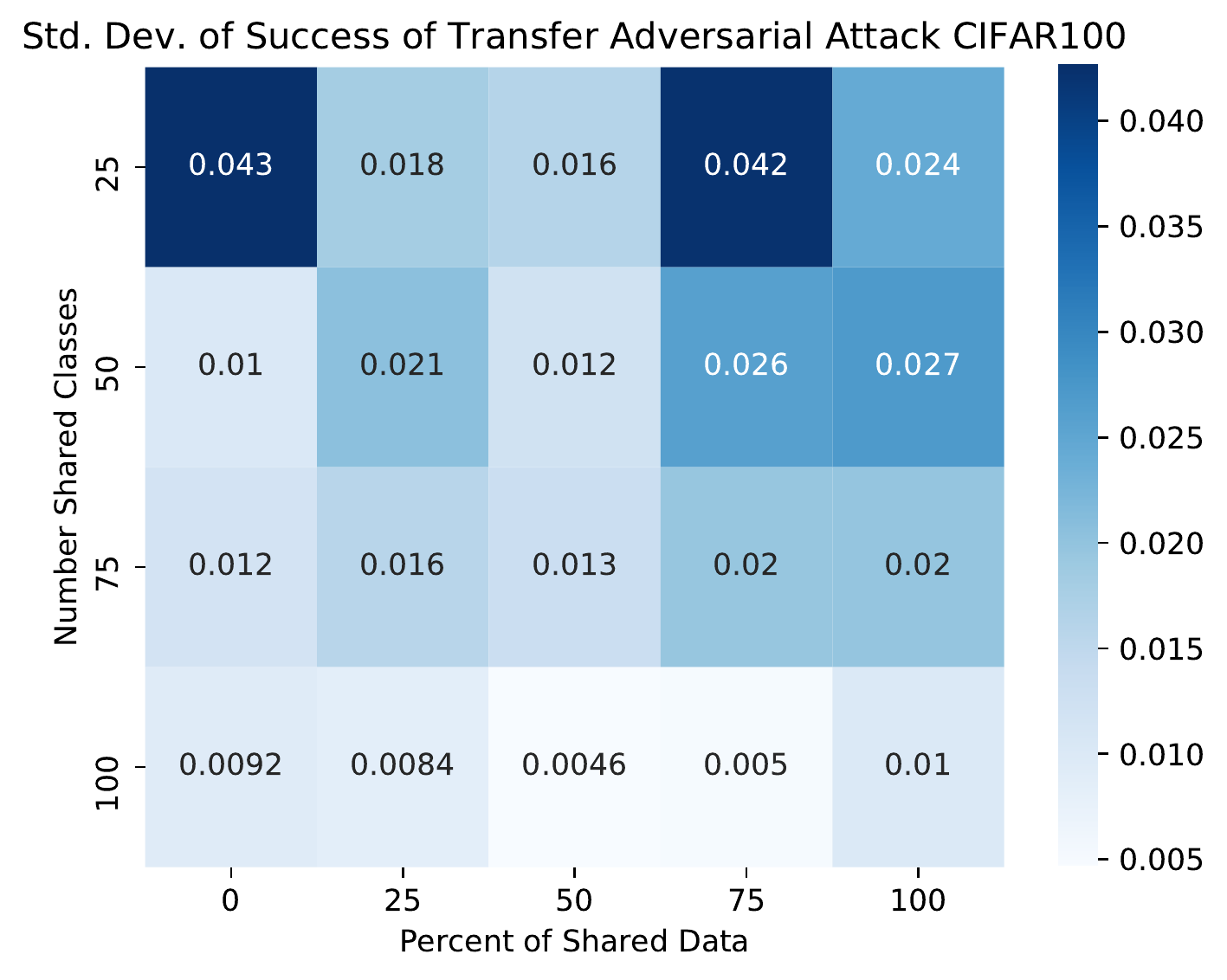}
\vspace{-2ex}
\caption{Standard deviation of success of transfer attack between 5 runs of each variation of shared classes and percentage of data shared for CIFAR100.}
\label{fig:CIFAR100CFNMTXSTD}
\end{figure}

CIFAR100 (\autoref{fig:CIFAR100CFNMTX}, \autoref{fig:CIFAR100CFNMTXSTD}) showcased more nuance to the threat model,
where the attacker's success improved as the number of shared classes \textit{decreased} with an unexpected negative correlation of -0.3068, with shared data now positive at 0.8118 (in contrast to being negative for the very similar CIFAR10). 
Given the intermix of adversarial attacks and out-of-distribution issues colliding, we do not yet have robust theory to explain or understand this. But the result 
 led us to examine the variance of the success between runs for smaller class overlaps. The variance in \autoref{fig:CIFAR100CFNMTXSTD} showed that smaller class overlap had increasingly high variance with less overlap. 
The higher variance as shared classes decreases is intrinsically important to the attacker: it means they have increased uncertainty of what any individual attacks performance will be. The variance is also not large enough to explain why fewer shared classes improve the attacker's average success rate. 

\begin{figure}[!h]
\centering
\vspace{1ex}
\includegraphics[width=1\columnwidth,scale=2.5]{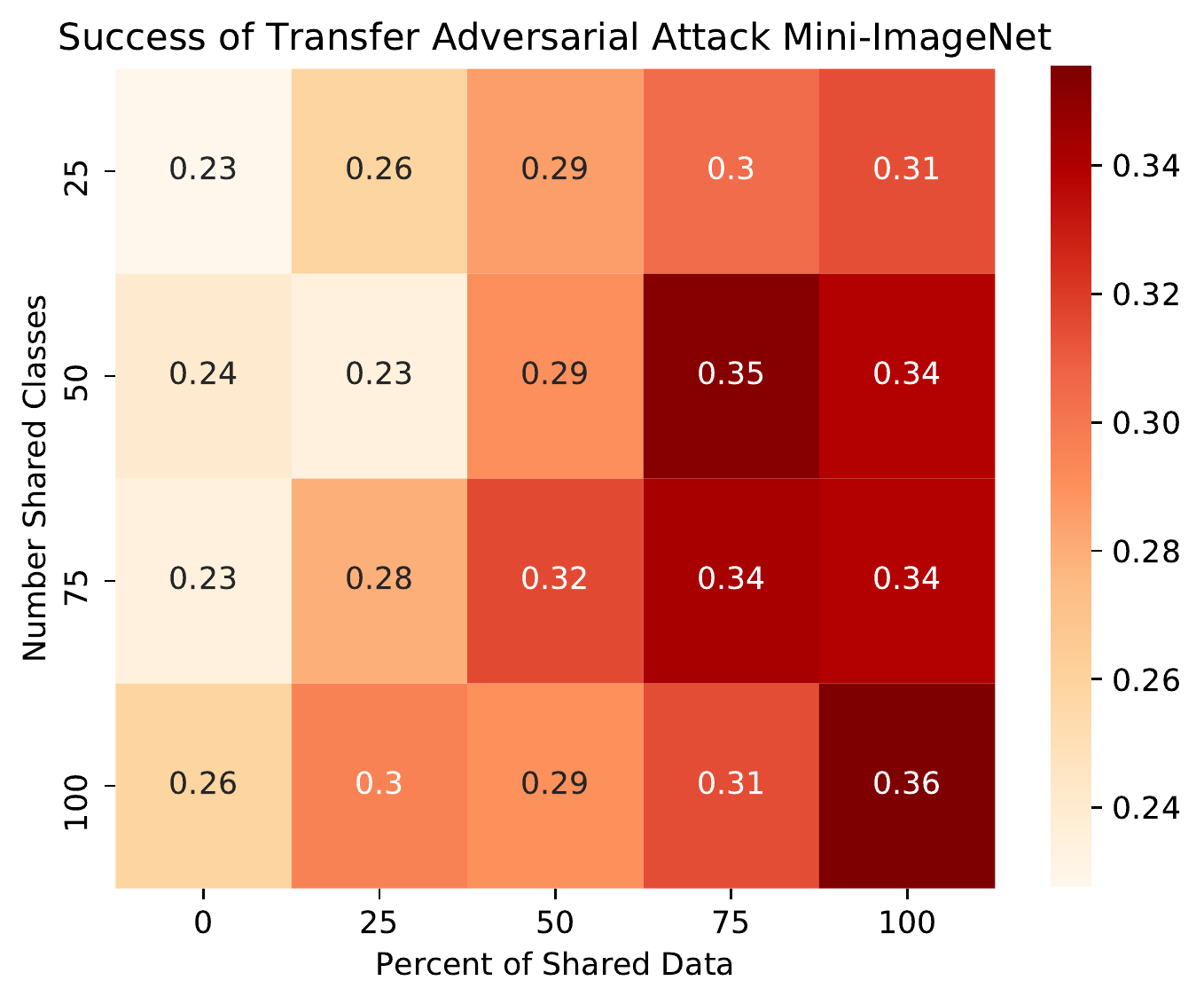}
\vspace{-2ex}
\caption{Success of transfer attack  dependent on variations of shared classes and percentage of data shared for Mini-ImageNet.}
\label{fig:MINIIMGNETSUCESS250}
\end{figure}

Finally we look at Mini ImageNet (\autoref{fig:MINIIMGNETSUCESS250}) which has 100 classes similar to CIFAR100. In this case, the data resolution  is larger \textit{(32 × 32 vs. 84 × 84)} and more diverse. 
At first glance it's result appear to match a priori expectations, shared classes have a positive correlation with attack success of 0.2170 and shared data a correlation of 0.8231. 
Again we observe the edge cases of intermediate shared percentages performing better than their extremes (e.g., 50\% shared classes and 75\% shared data has a 35\% transfer success, higher than \textit{all} neighboring values).

\begin{figure}[!h]
\centering
\vspace{1ex}
\includegraphics[width=1\columnwidth,scale=2.5]{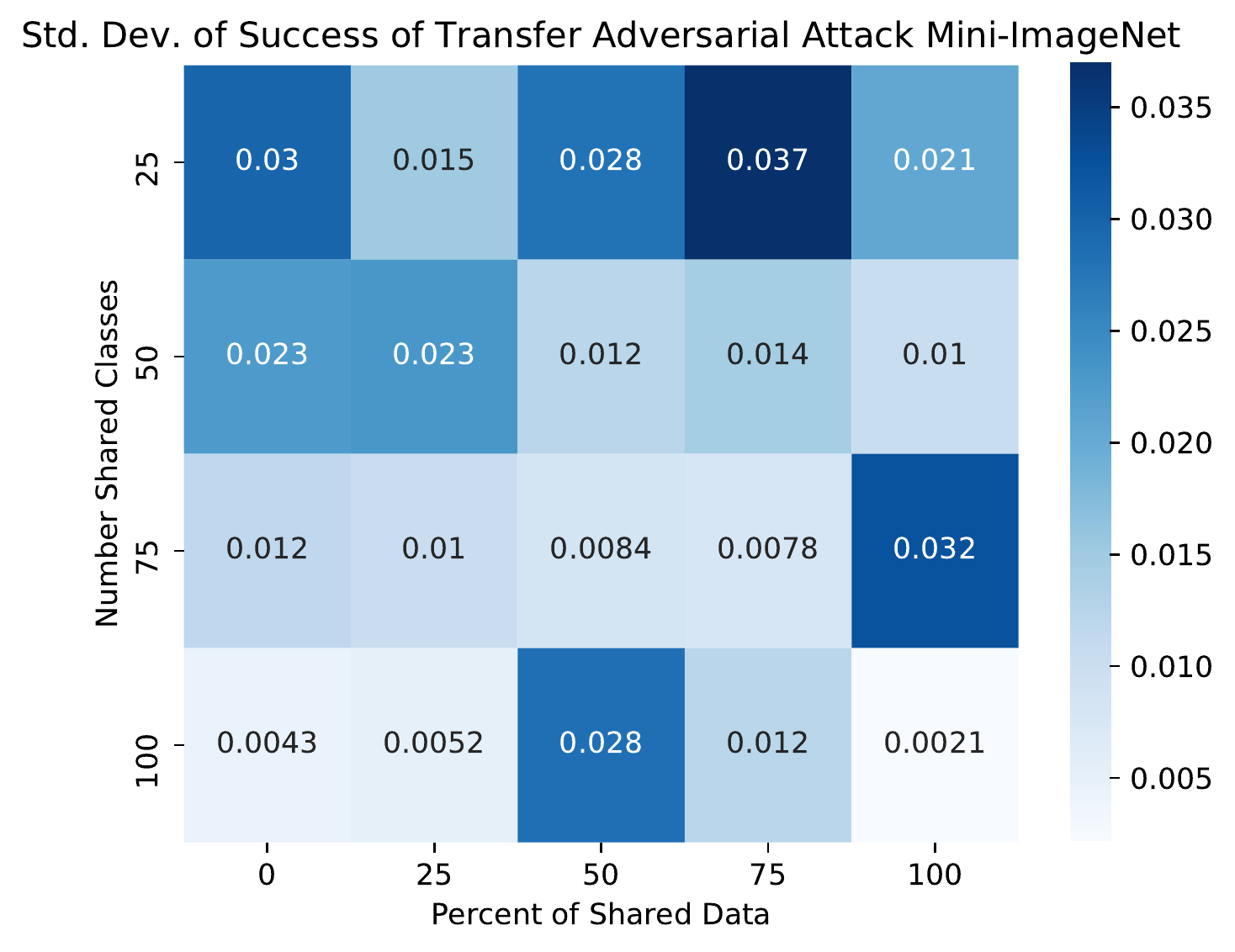}
\vspace{-2ex}
\caption{Standard deviation of transfer attack  dependent on variations of shared classes and percentage of data shared for Mini-ImageNet.}
\label{fig:MINIIMGNETSUCESS250_STD}
\end{figure}

The variance of Mini-ImageNet results (see \autoref{fig:MINIIMGNETSUCESS250_STD}) matches the same trend in variance from CIFAR-100:
There is
more variance with lower overlap in both number classes. 
This is the only consistent and ``intuitive'' behavior that occurs in this threat model, but is a pyrrhic win at best for the attacker. The higher variance may lead to greater success, but the attacker cannot rely on this to occur, and so must account for the variance in results and select from the lower tail of the distribution to make a ``safe'' decision about whether or not their attack will be successful. 

For the results in this section, we start to gain an understanding of the behavior dynamics within this threat model. While we are unable to observe the intuition of a linear relationship for both factors, we are able to identify the nuance of variance that makes this threat model unique and challenging. The cause of these non-intuitive behaviors is not known, and a matter of future works given the intermix of adversarial and out-of-distribution issues that blend together. Independently both issues have little existing theory, and we are aware of none that exists for the union of these problems we study. 

\begin{figure}[!h]
\centering
\vspace{1ex}
\includegraphics[width=1\columnwidth,scale=2.5]{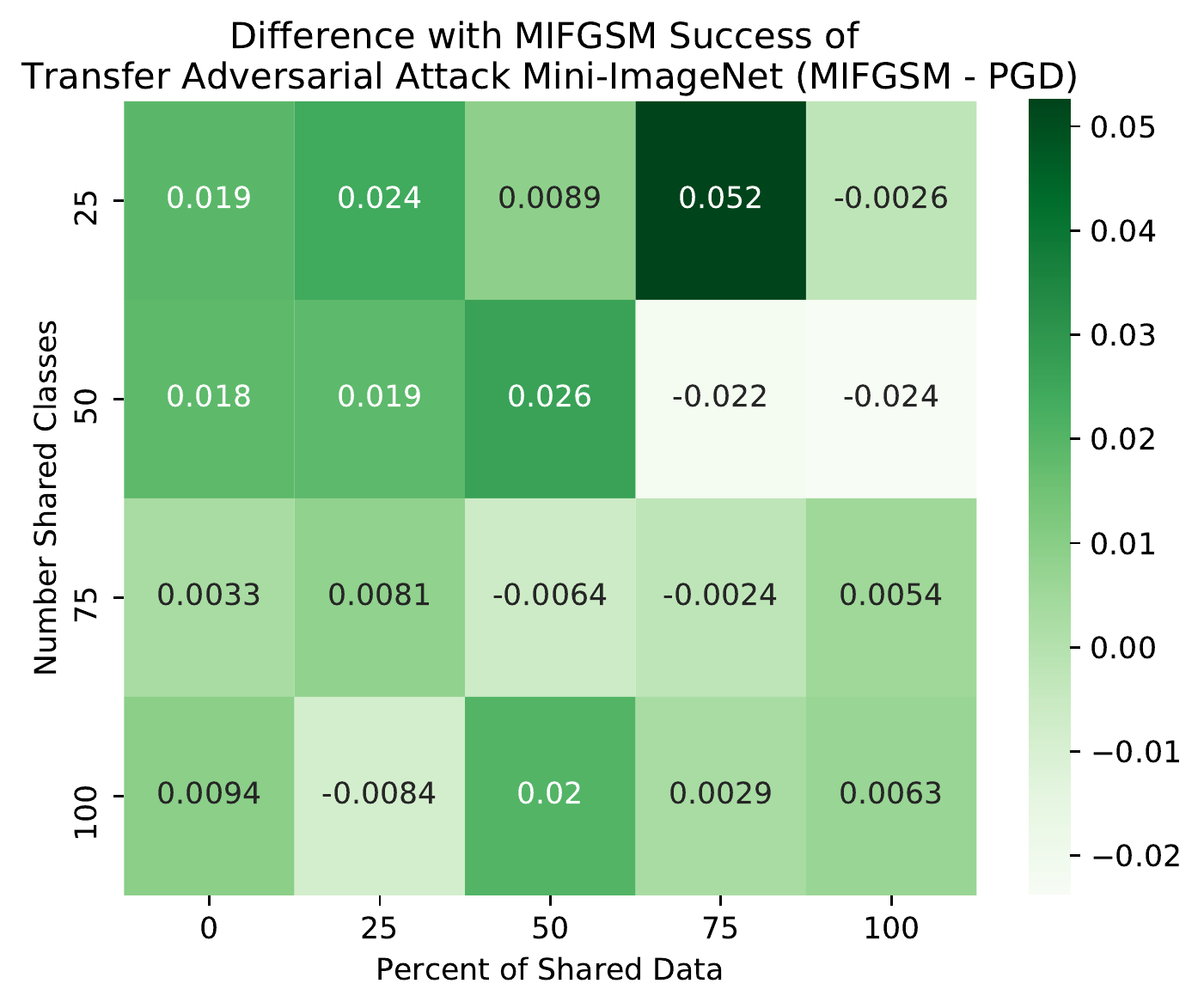}
\vspace{-2ex}
\caption{Difference between MI-FGSM attack and PGD attack in terms of success in attack. Lighter means PGD had more success and darker means MI-FGSM had more success.}
\label{fig:MINIIMGNETSUCESS250_MIFGSM}
\end{figure}
We also examine work crafting attack objectives meant to improve transfer to a victim model. We evaluated the standard PDG attack against Momentum Iterative Fast Gradient Sign Method (MIFGSM) \cite{dong2018boosting}. We found within our comparative results on mini-ImageNet (see \autoref{fig:MINIIMGNETSUCESS250_MIFGSM}) that the method offered small success percentages over PGD for transfer in some cases. We defer to future work for examining transfer variants of attacks and their performance within this framework.

\subsection{Adversarial Hardening}

When training both the surrogate model and the victim model with adversarial examples included we saw two main takeaways. 1) the  vulnerability of the hardened model in most cases was not as robust against attacks as the vanilla trained model. 2) we find that there is an interesting behavior profile in which the hardening is most effective with class overlap opposed to data overlap. 

\begin{figure}[!h]
\centering
\vspace{1ex}
\includegraphics[width=1\columnwidth,scale=2.5]{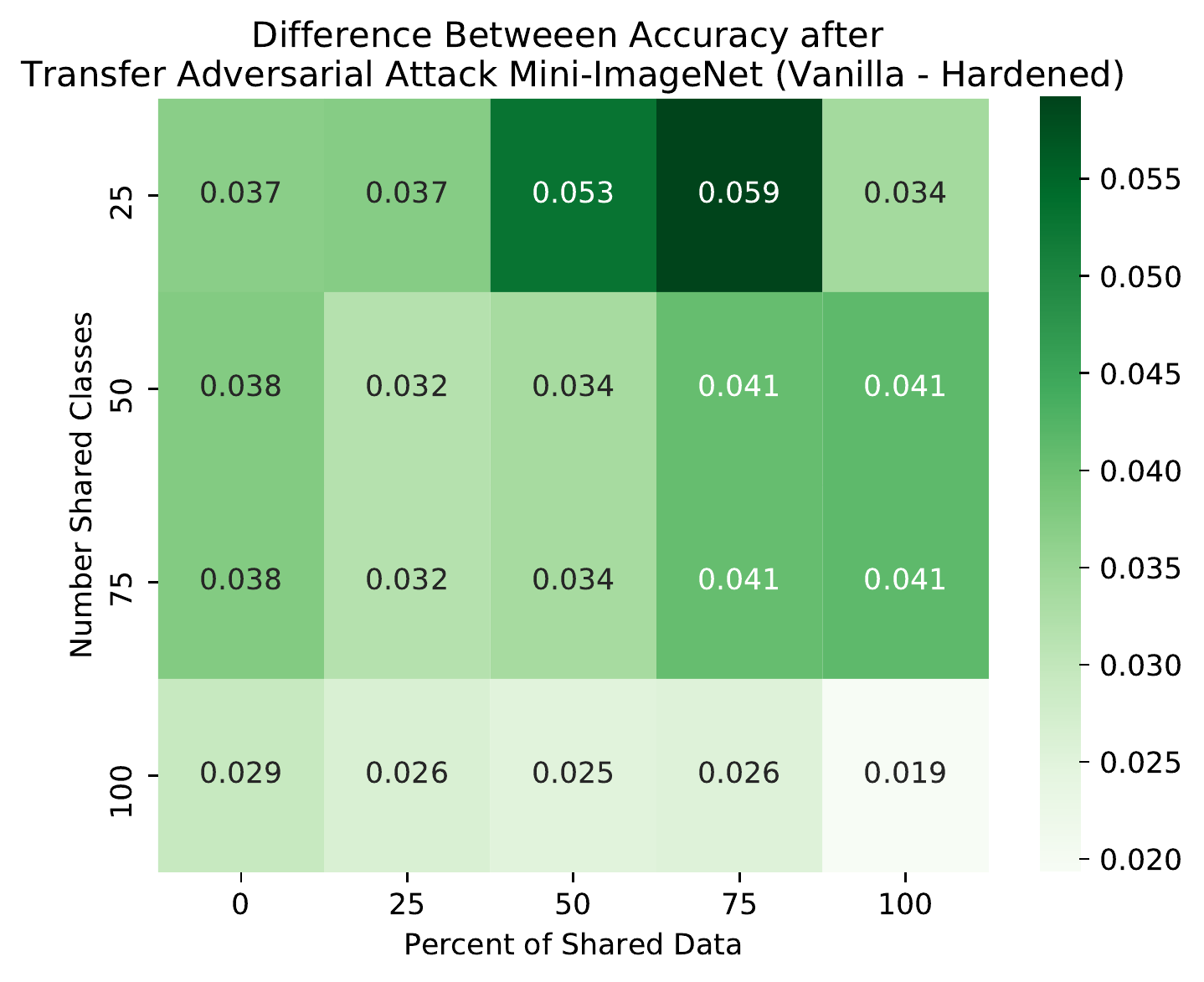}
\vspace{-2ex}
\caption{The difference between vanilla and hardened models' accuracy. Overall the hardened model has lower accuracy in all cases with the least being complete overlap, the only context adversarial hardening has been studied.}
\label{fig:MINIIMGNETDIFFHARDNED}
\end{figure}

\begin{figure}[!h]
\centering
\vspace{1ex}
\includegraphics[width=1\columnwidth,scale=1]{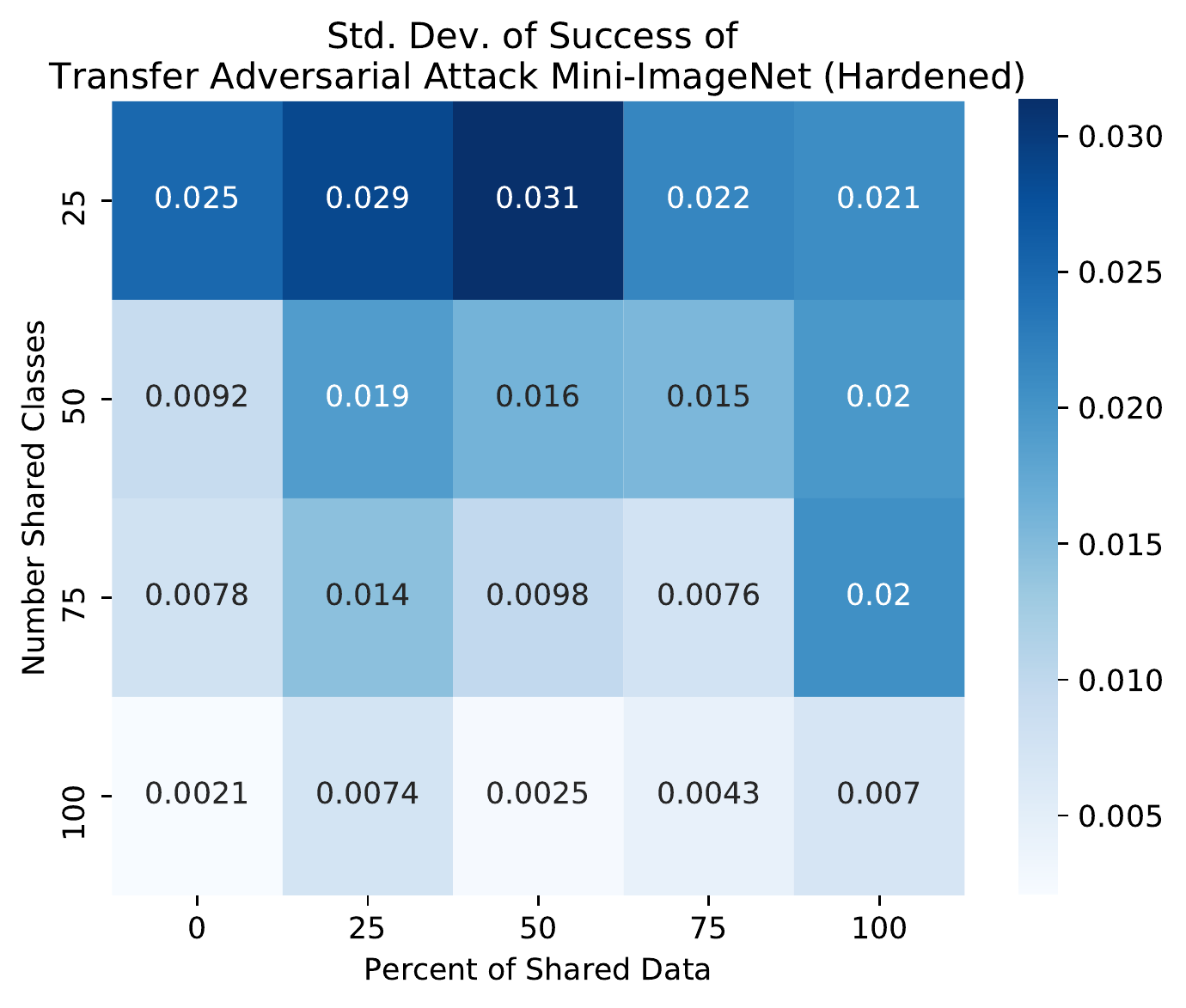}
\vspace{-2ex}
\caption{Standard deviation between success of runs of transfer attack on adversarial hardened model. We again see the same level of variation as with the non-hardened model.}
\label{fig:MINIIMGNETADVHARDSTDV}
\end{figure}

\begin{figure}[!h]
\centering
\vspace{1ex}
\includegraphics[width=1\columnwidth,scale=2.5]{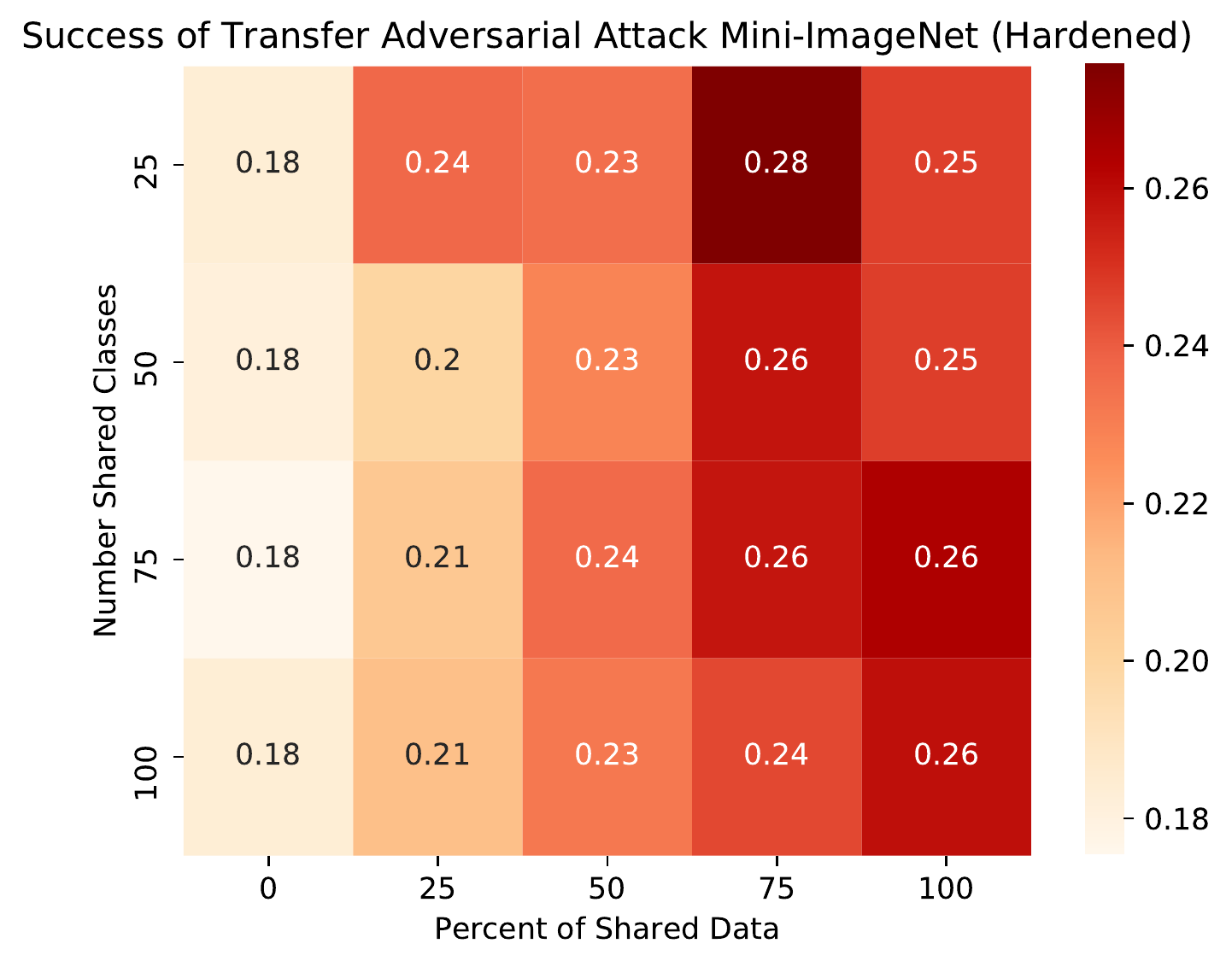}
\vspace{-2ex}
\caption{Transfer success with adversarial hardening during training. Overall we saw lower accuracy and then less success. We chose to include the differences in accuracies in the paper to show that the cost of robustness only drastically assisted with the assumption of total overlap.}
\label{fig:MINIIMGNETADVHARD}
\end{figure}

We find it valuable to frame the performance within the context of the difference between the vanilla model and hardened model's accuracies when under attack, as shown in  \autoref{fig:MINIIMGNETDIFFHARDNED}. This distinction sheds light on the trade off between robustness and accuracy that previous work faced. The correlation coefficients for these values on the Mini-ImageNet dataset for number of shared classes is -0.5537. We show both the success (\autoref{fig:MINIIMGNETADVHARD}) and standard deviation between runs (\autoref{fig:MINIIMGNETADVHARDSTDV}) when hardening is applied. 
This indicates that as the attacker and victim share fewer classes, the adversarial training becomes increasingly \textit{detrimental} to the victim's goal of correct predictions while under attack, dropping by as much as 5.9 percentage points. 
The correlation with respect to shared data is only 0.0136, indicating that having the exact same data or no shared data is of comparatively little impact on the outcome. 

The normal behavior is that adversarial training is purely beneficial, and usually the state-of-the-art defense against attacks (if within your computational constraints). However, implicit within the framework of adversarial training is that the attacker and defender have the exact same classes, the same data distribution, and thus the same manifold on which the data and attacks lie (though potentially different estimates of that manifold). Our results indicate that adversarial training can thus ``overfit'' to the specifics of the current data. When attacks are generated from a highly differentiated manifold (i.e., low \% shared classes) they exhibit behavior and values far outside what the defender could have generated using adversarial training, and thus, does not generalize to the attacked inputs. This places even greater importance on research into provably secure defenses for the defender's confidence, but such study is outside the scope of our current work (and compute resources).

\subsection{Masked PGD Attack}

\begin{figure}[!h]
\centering
\vspace{1ex}
\includegraphics[width=1\columnwidth,scale=2.5]{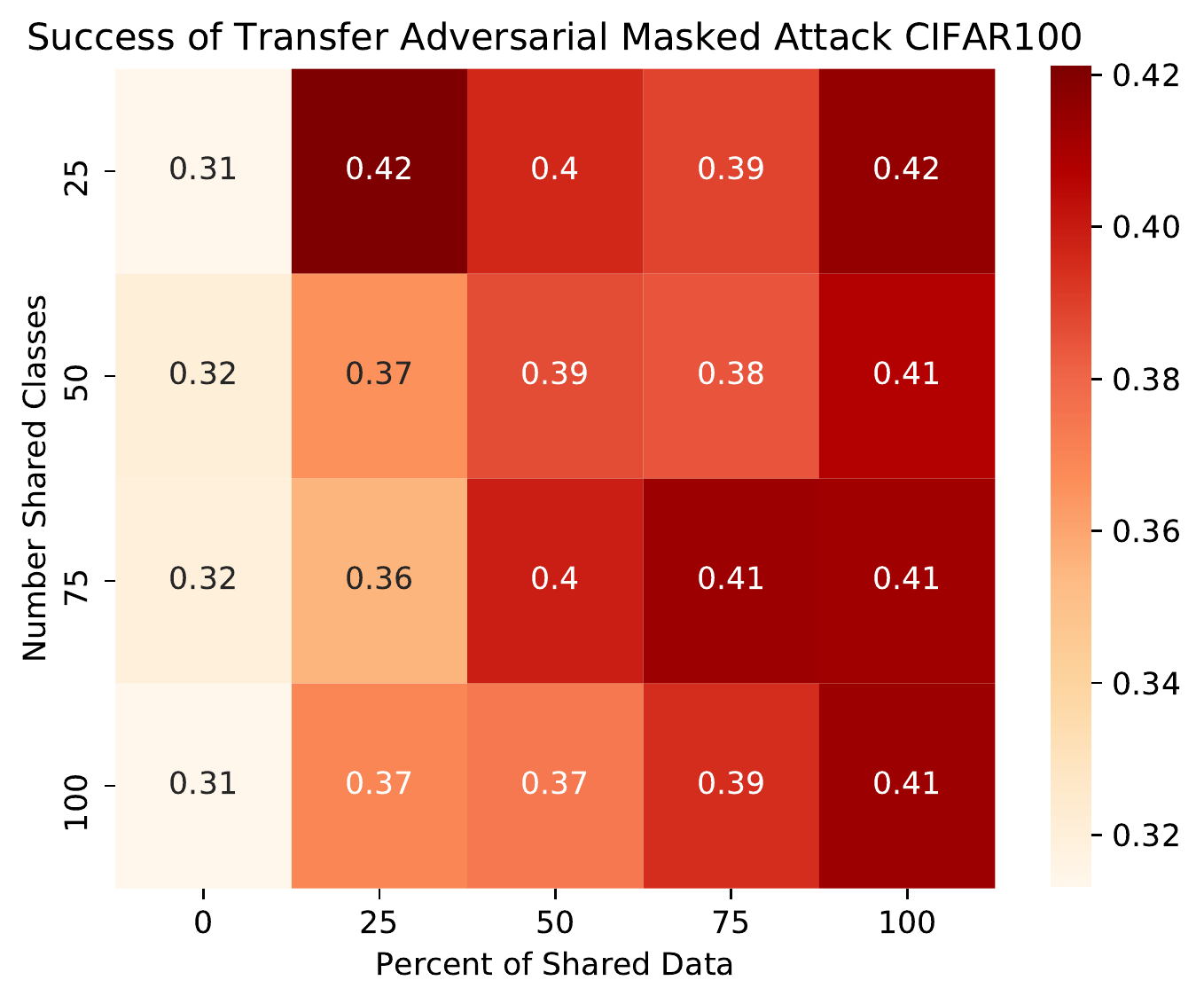}
\vspace{-2ex}
\caption{Success of masked transfer attack  dependent on variations of shared classes and percentage of data shared for CIFAR100.}
\label{fig:CIFAR10CFNMTXMASKED}
\end{figure}

\begin{figure}[!h]
\centering
\vspace{1ex}
\includegraphics[width=1\columnwidth,scale=2.5]{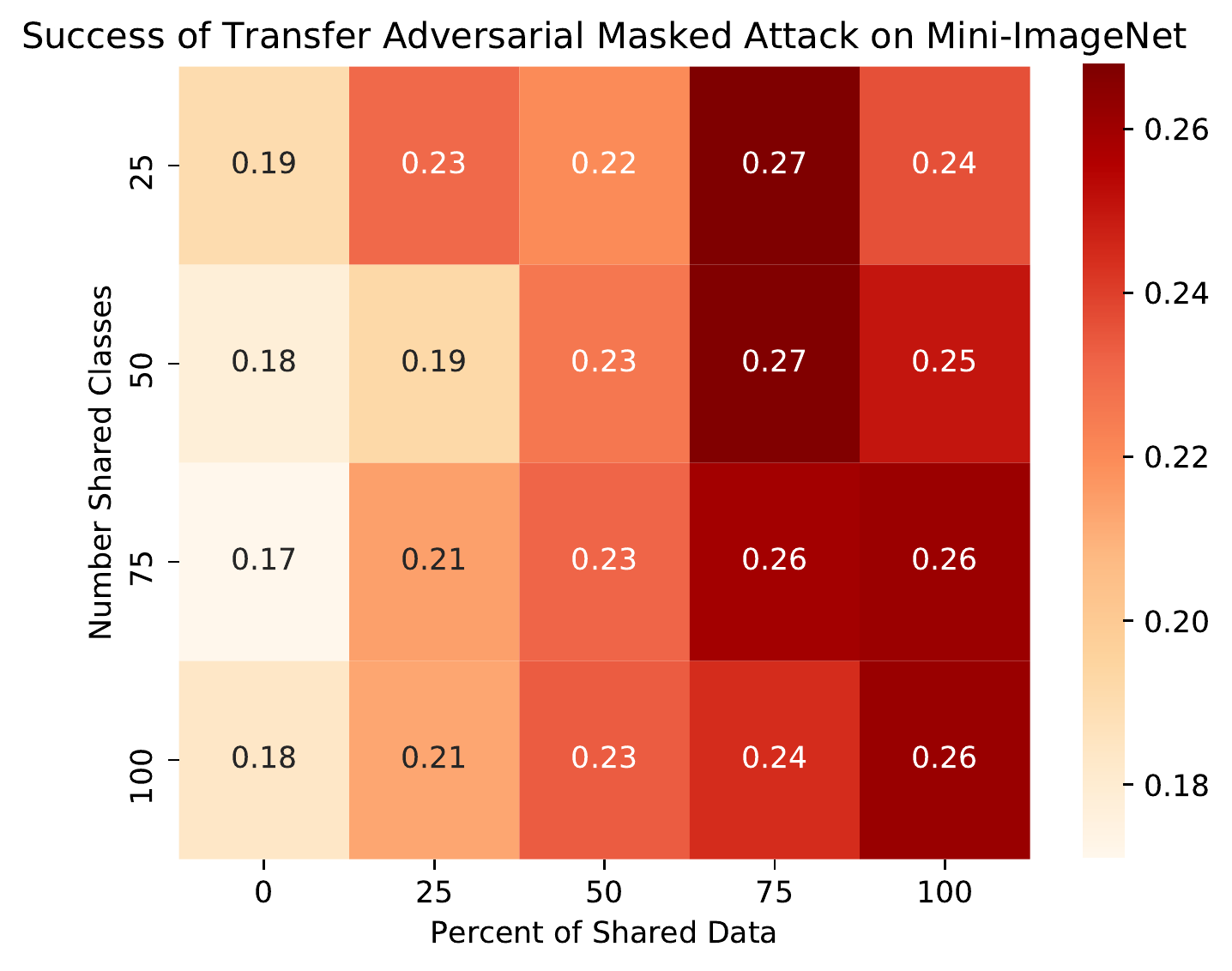}
\vspace{-2ex}
\caption{Attack success with masked PGD attack dependent on variations of shared classes and percentage of data shared for Mini-ImageNet.}
\label{fig:MINIIMGNETMASKED}
\end{figure}

\begin{figure}[!h]
\centering
\vspace{1ex}
\includegraphics[width=1\columnwidth,scale=2.5]{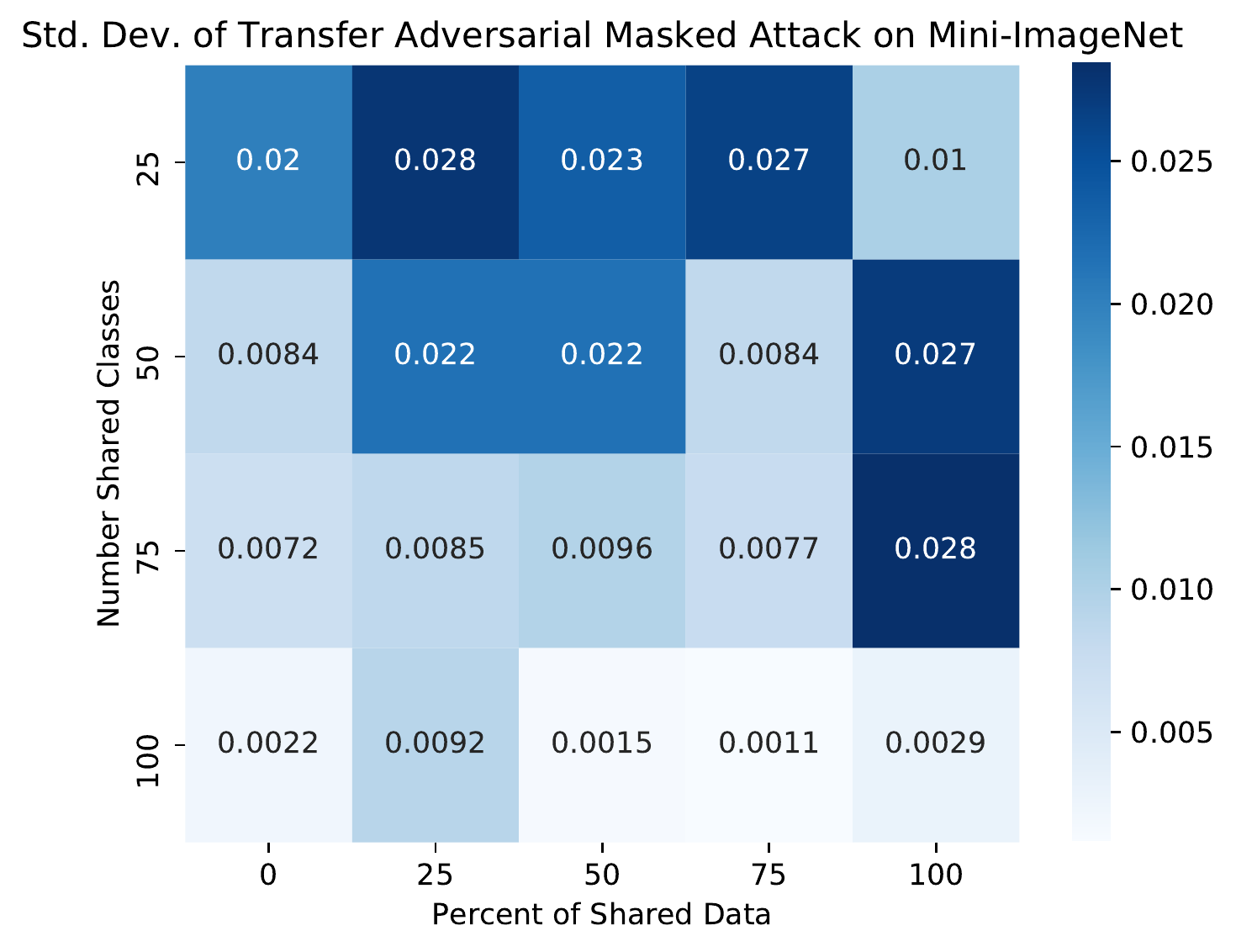}
\vspace{-2ex}
\caption{Standard deviation of transfer masked-based attack dependent on variations of shared classes and percentage of data shared for Mini-ImageNet.}
\label{fig:MINIIMAGENETMASKEDSTDV}
\end{figure}

Our results thus far have demonstrated that current intuitions and expectations on the behavior of adversarial attacks break down when there is a discordance between attacker's surrogate and the victim models. This makes reasoning for both perspectives difficult, and computationally demanding. Instead, we show that our Masked PGD attack can provide more stable estimates of attacker success that almost returns to the expected behaviors. The attack success rate is shown in \autoref{fig:MINIIMGNETMASKED}, where the variance with respect to the number of shared classes is almost entirely eliminated, and what little remains follows expectation that fewer shared classes results in a lower transfer success rate. We also see the exact same behavior profile occur on CIFAR100, as shown in \autoref{fig:CIFAR10CFNMTXMASKED}, and that the variance in results match continue to remain low when all classes are shared (\autoref{fig:MINIIMAGENETMASKEDSTDV}). We are unable to compare exhaustively or perform as many iterations, but we also see the same general trend on the full ImageNet corpus which is shown in \autoref{sec:appendix_imagenet}, which has greater variance due to the extreme computational expense which prevents us from doing more than one trial, where our CIFAR100 and Mini-ImageNet results are the average from multiple trials. 

We note that the extreme expense for these experiments is caused by needing multiple runs of multiple difference combinations of shared class and data overlap percentages. In practice, because our Masked PGD attack restores behavior expectations, the attacker or defender can simply perform the Masked PGD attack for only the case of low shared classes and 0\% shared data. This provides attack and defender with a reasonable and reliable lower-bound for the attacker's probability of success without the $\geq 20\times$ cost of evaluation for each possible combination. 

Applying our Masked PGD during adversarial training is not as trivial, and in our initial experiments introduced significant decreases in stability and accuracy, as well as out-sized computational cost (the number of PGD iterations, multiplied by the number of masked iterations). Our results thus provide a means for better estimating transfer success in our threat model, but not yet a mechanism for improved defense/robustness. This is critical for future work as our results will show that adversarial training \textit{reduces} defender robustness in this threat model.

\section{Initial Image Net Results} \label{sec:appendix_imagenet}

The exploration of our results on an ImageNet scale is of extreme cost. 
Further understanding the results at the scale of ImageNet would also be valuable beyond the current contributions.
We choose to document the initial results which are incomplete, necessarily so because they are too expensive for us to complete without external assistance. Our hope is that these incomplete results will help to illustrate the high cost for the adversary in attacking and understanding their likelihood of success as the number of classes and data increases, but also the open problem of estimating these properties for both the adversary and defender for future work. 

\begin{figure}[!h]
\centering
\vspace{1ex}
\includegraphics[width=1\columnwidth,scale=2.5]{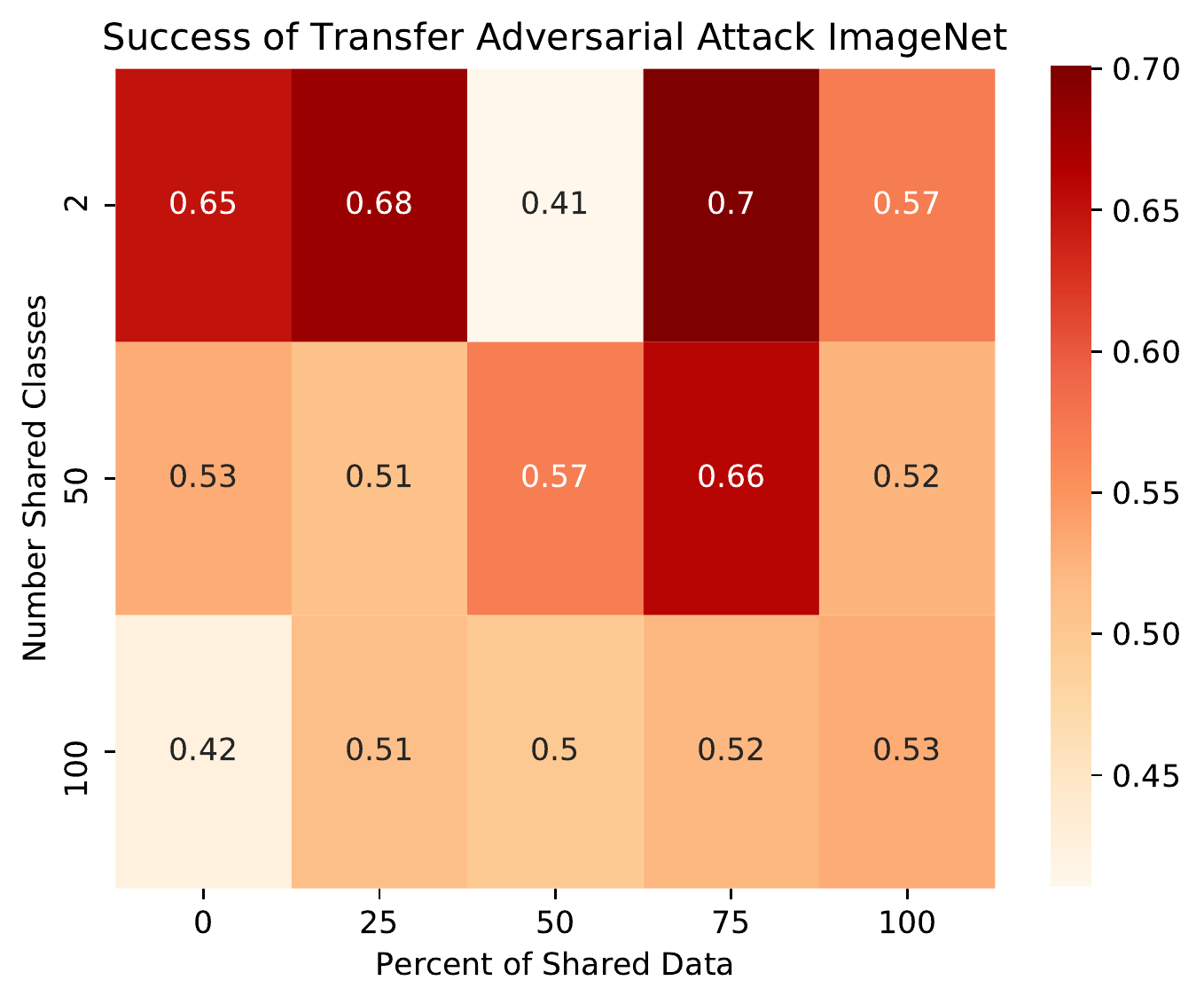}
\vspace{-2ex}
\caption{Success of transfer attack  dependent on variations of shared classes and percentage of data shared for partial ImageNet.}
\label{fig:IMGNET}
\end{figure}

\begin{figure}[!h]
\centering
\vspace{1ex}
\includegraphics[width=1\columnwidth,scale=2.5]{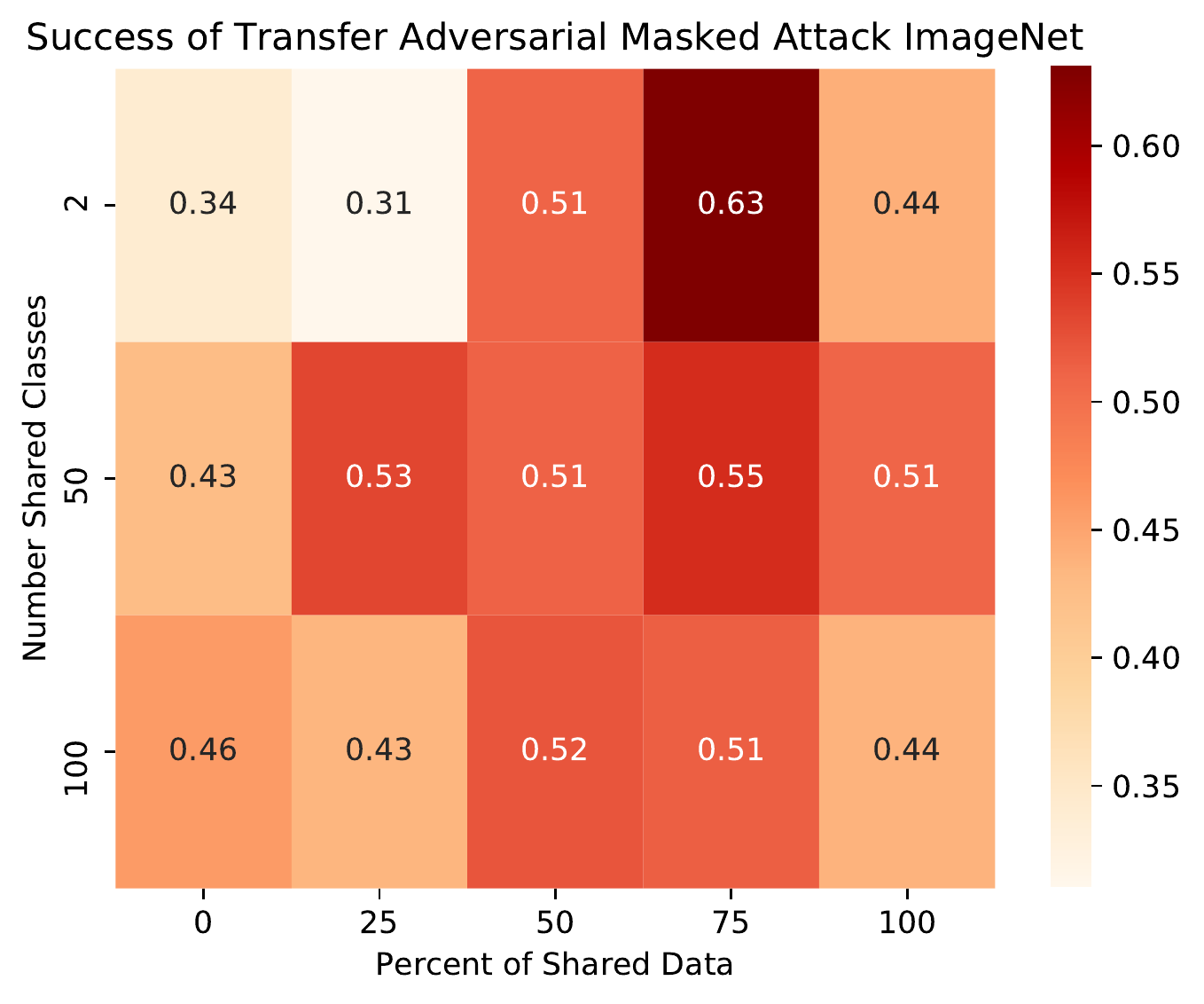}
\vspace{-2ex}
\caption{Success of masked transfer attack  dependent on variations of shared classes and percentage of data shared for partial ImageNet.}
\label{fig:IMGNETMASKED}
\end{figure}

Initial experiments were run on ImageNet with 2, 50, and 100 shared classes (2\%, 5\%, and 10\% of all ImageNet classes) each with 0\%, 25\%, 50\%, 75\%, and 100\% shared data percentages. Comparing to our experiments (\autoref{fig:IMGNET}, \autoref{fig:IMGNETMASKED}) with CIFAR100, we saw large variance between runs with lower number of shared classes. This may be a factor within these initial results as well as we saw the most variance at 2\% of the shared classes in CIFAR100.  

These experiments were run parallel on a DGX v100 server with four 32GB GPUs. With the runs for 2, 50, and 100 shared classes taking 76, 144, and 164 GPU hours respectively. The increase is due to the number of data points that are removed with cutting the number of classes to create equal training data amount per model. For the masked attack that requires more compute, the GPU hours were 396, 476, and 584.

These numbers account for single runs in which the variance cannot be measured like previous analysis on smaller datasets. Due to this, we can assume that each run would need to be run 5 or more times to analyze this variance. Based on the early averages, we estimate in GPU hours this would take an estimated 2,200 GPU hours (assuming linear relationship) to complete one round of 5 class step and 5 percentage step experiments. With running each iteration 5 times to measure variance, it would be 11,100 GPU hours. With the masked version of the attack we estimate 5,980 GPU hours based on early results with a total of 29,900 GPU hours to analyze variance between runs.   

We foresee scaling these experiments on a large dataset such as ImageNet to be a considerable engineering contribution in itself. Crucially, introducing the security analysis paradigm of the intersection between shared classes and shared data points itself shows promising results on smaller datasets.

\section{Conclusion} \label{sec:conclusion}

We have elucidated a data-centric threat model where the adversary cannot query the victim, but must still transfer evasion attacks using a surrogate. Many prior works have implicitly used this model, with the unrealistic setting of the adversary knowing all the exact data used by the victim. This model is applicable to many real-world situations, and we have shown for the first time the asperity of attack success rate as the amount of shared data and classes vary. 
Attack success rate and correlations change sign depending on dataset used, and adversarial training can actually reduce defender success, making it difficult for attacker and defender to estimate overall success. Our new Masked PGD attack removes the variance with respect to shared classes, and restores an intuitive behavior to results across all tested datasets. This allows the attacker to confidently estimate attack success rate. 

Considerable work still remains in fully and more accurately modeling real-world threat models under imperfect information and observation. In particular our work has taken a first step by sub-sampling existing classes, but more complex partial intersection of victim vs surrogate classes is likely in real world scenarios. The large computational cost to study these scenarios due to the intrinsically stochastic nature of class overlap also needs addressing. In particular this impacts the feasibility of adversarial training, which loses significant utility in our testing. 

{\small
\bibliographystyle{ieee_fullname}
\bibliography{Mendeley-Andre,references-raff,references}

\begin{thebibliography}{10}\itemsep=-1pt

\bibitem{Anderson2018}
Hyrum~S. Anderson and Phil Roth.
\newblock {EMBER: An Open Dataset for Training Static PE Malware Machine
  Learning Models}.
\newblock {\em ArXiv e-prints}, 2018.

\bibitem{Arslan2017}
Bilgehan Arslan, Mehtap Ulker, and Seref Sagiroglu.
\newblock {Machine Learning Methods Used in Evaluations of Secure Biometric
  System Components}.
\newblock In {\em 2017 16th IEEE International Conference on Machine Learning
  and Applications (ICMLA)}, pages 448--453. IEEE, 12 2017.

\bibitem{Ashraf2009}
Ahmed~Bilal Ashraf, Simon Lucey, Jeffrey~F. Cohn, Tsuhan Chen, Zara Ambadar,
  Kenneth~M. Prkachin, and Patricia~E. Solomon.
\newblock {The painful face – Pain expression recognition using active
  appearance models}.
\newblock {\em Image and Vision Computing}, 27(12):1788--1796, 11 2009.

\bibitem{Athalye2018}
Anish Athalye, Nicholas Carlini, and David Wagner.
\newblock {Obfuscated Gradients Give a False Sense of Security: Circumventing
  Defenses to Adversarial Examples}.
\newblock In {\em International Conference on Machine Learning (ICML)}, 2018.

\bibitem{Athalye2017}
Anish Athalye, Logan Engstrom, Andrew Ilyas, and Kevin Kwok.
\newblock {Synthesizing Robust Adversarial Examples}.
\newblock {\em ArXiv e-prints}, 2017.

\bibitem{Biggio2014}
Battista Biggio, Giorgio Fumera, and Fabio Roli.
\newblock {Security evaluation of pattern classifiers under attack}.
\newblock {\em IEEE Transactions on Knowledge and Data Engineering},
  26(4):984--996, 2014.

\bibitem{Biggio2017}
Battista Biggio and Fabio Roli.
\newblock {Wild patterns: Ten years after the rise of adversarial machine
  learning}.
\newblock {\em Pattern Recognition}, 84:317--331, 12 2018.

\bibitem{Broome2019}
Sofia Broome, Karina~Bech Gleerup, Pia~Haubro Andersen, and Hedvig Kjellstrom.
\newblock {Dynamics Are Important for the Recognition of Equine Pain in Video}.
\newblock In {\em 2019 IEEE/CVF Conference on Computer Vision and Pattern
  Recognition (CVPR)}, pages 12659--12668. IEEE, 6 2019.

\bibitem{Carlini:2017:AEE:3128572.3140444}
Nicholas Carlini and David Wagner.
\newblock {Adversarial Examples Are Not Easily Detected: Bypassing Ten
  Detection Methods}.
\newblock In {\em Proceedings of the 10th ACM Workshop on Artificial
  Intelligence and Security}, AISec '17, pages 3--14, New York, NY, USA, 2017.
  ACM.

\bibitem{10.1145/3394486.3403368}
Wei-Cheng Chang, Hsiang-Fu Yu, Kai Zhong, Yiming Yang, and Inderjit~S Dhillon.
\newblock {Taming Pretrained Transformers for Extreme Multi-Label Text
  Classification}.
\newblock In {\em Proceedings of the 26th ACM SIGKDD International Conference
  on Knowledge Discovery {\&}amp; Data Mining}, KDD '20, page 3163–3171, New
  York, NY, USA, 2020. Association for Computing Machinery.

\bibitem{cheng2019improving}
Shuyu Cheng, Yinpeng Dong, Tianyu Pang, Hang Su, and Jun Zhu.
\newblock Improving black-box adversarial attacks with a transfer-based prior.
\newblock {\em arXiv preprint arXiv:1906.06919}, 2019.

\bibitem{Demontis2018}
Ambra Demontis, Marco Melis, Maura Pintor, Matthew Jagielski, Battista Biggio,
  Alina Oprea, Cristina Nita-Rotaru, and Fabio Roli.
\newblock {Why Do Adversarial Attacks Transfer? Explaining Transferability of
  Evasion and Poisoning Attacks}.
\newblock {\em ArXiv e-prints}, pages 26--28, 2018.

\bibitem{236234}
Ambra Demontis, Marco Melis, Maura Pintor, Matthew Jagielski, Battista Biggio,
  Alina Oprea, Cristina Nita-Rotaru, and Fabio Roli.
\newblock {Why Do Adversarial Attacks Transfer? Explaining Transferability of
  Evasion and Poisoning Attacks}.
\newblock In {\em 28th USENIX Security Symposium (USENIX Security 19)}, pages
  321--338, Santa Clara, CA, 8 2019. {\{}USENIX{\}} Association.

\bibitem{dong2018boosting}
Yinpeng Dong, Fangzhou Liao, Tianyu Pang, Hang Su, Jun Zhu, Xiaolin Hu, and
  Jianguo Li.
\newblock Boosting adversarial attacks with momentum.
\newblock In {\em Proceedings of the IEEE conference on computer vision and
  pattern recognition}, pages 9185--9193, 2018.

\bibitem{8953425}
Y Dong, T Pang, H Su, and J Zhu.
\newblock {Evading Defenses to Transferable Adversarial Examples by
  Translation-Invariant Attacks}.
\newblock In {\em 2019 IEEE/CVF Conference on Computer Vision and Pattern
  Recognition (CVPR)}, pages 4307--4316, 6 2019.

\bibitem{dong2019evading}
Yinpeng Dong, Tianyu Pang, Hang Su, and Jun Zhu.
\newblock Evading defenses to transferable adversarial examples by
  translation-invariant attacks.
\newblock In {\em Proceedings of the IEEE/CVF Conference on Computer Vision and
  Pattern Recognition}, pages 4312--4321, 2019.

\bibitem{Egele2017}
Manuel Egele, T Scholte, E Kirda, and Santa Barbara.
\newblock {A Survey On Automated Dynamic Malware Analysis Evasion and
  Counter-Evasion}.
\newblock In {\em Proceedings of Reversing and Offensive-oriented Trends
  Symposium}, 2017.

\bibitem{Finlayson}
Samuel~G Finlayson, Isaac~S Kohane, and Andrew~L Beam.
\newblock {Adversarial Attacks Against Medical Deep Learning Systems}.
\newblock 2018.

\bibitem{Hammal2012}
Zakia Hammal and Jeffrey~F. Cohn.
\newblock {Automatic detection of pain intensity}.
\newblock In {\em Proceedings of the 14th ACM international conference on
  Multimodal interaction - ICMI '12}, page~47, New York, New York, USA, 2012.
  ACM Press.

\bibitem{he15deepresidual}
Kaiming He, Xiangyu Zhang, Shaoqing Ren, and Jian Sun.
\newblock {Deep Residual Learning for Image Recognition}.
\newblock In {\em IEEE Conference on Computer Vision and Pattern Recognition
  (CVPR)}, 2015.

\bibitem{10.1145/3289600.3290979}
Himanshu Jain, Venkatesh Balasubramanian, Bhanu Chunduri, and Manik Varma.
\newblock {Slice: Scalable Linear Extreme Classifiers Trained on 100 Million
  Labels for Related Searches}.
\newblock In {\em Proceedings of the Twelfth ACM International Conference on
  Web Search and Data Mining}, WSDM '19, page 528–536, New York, NY, USA,
  2019. Association for Computing Machinery.

\bibitem{Kariyappa2019}
Sanjay Kariyappa and Moinuddin~K. Qureshi.
\newblock {Improving Adversarial Robustness of Ensembles with Diversity
  Training}.
\newblock {\em arXiv}, 2019.

\bibitem{Kingma2015}
Diederik~P Kingma and Jimmy~Lei Ba.
\newblock {Adam: A Method for Stochastic Optimization}.
\newblock In {\em International Conference On Learning Representations}, 2015.

\bibitem{Kolosnjaji2018AdversarialExecutables}
Bojan Kolosnjaji, Ambra Demontis, Battista Biggio, Davide Maiorca, Giorgio
  Giacinto, Claudia Eckert, and Fabio Roli.
\newblock {Adversarial malware binaries: Evading deep learning for malware
  detection in executables}.
\newblock {\em European Signal Processing Conference}, 2018-Septe:533--537,
  2018.

\bibitem{Kurakin2017}
Alexey Kurakin, Ian Goodfellow, and Samy Bengio.
\newblock {Adversarial Machine Learning at Scale}.
\newblock In {\em International Conference on Learning Representations (ICLR)},
  2017.

\bibitem{Lucey2009}
Patrick Lucey, Jeffrey Cohn, Simon Lucey, Iain Matthews, Sridha Sridharan, and
  Kenneth~M. Prkachin.
\newblock {Automatically detecting pain using facial actions}.
\newblock In {\em 2009 3rd International Conference on Affective Computing and
  Intelligent Interaction and Workshops}, pages 1--8. IEEE, 9 2009.

\bibitem{Madry2018}
Aleksander Madry, Aleksandar Makelov, Ludwig Schmidt, Dimitris Tsipras, and
  Adrian Vladu.
\newblock {Towards Deep Learning Models Resistant to Adversarial Attacks}.
\newblock In {\em International Conference on Learning Representations (ICLR)},
  2018.

\bibitem{Martinez2017}
Daniel~Lopez Martinez, Ognjen Rudovic, and Rosalind Picard.
\newblock {Personalized Automatic Estimation of Self-Reported Pain Intensity
  from Facial Expressions}.
\newblock In {\em 2017 IEEE Conference on Computer Vision and Pattern
  Recognition Workshops (CVPRW)}, pages 2318--2327. IEEE, 7 2017.

\bibitem{Nguyen2018AdversarialRegularization}
Andre~T Nguyen and Edward Raff.
\newblock {Adversarial Attacks, Regression, and Numerical Stability
  Regularization}.
\newblock In {\em AAAI 2019 Workshop on Engineering Dependable and Secure
  Machine Learning Systems}, 2018.

\bibitem{Papernot2016}
Nicolas Papernot, Patrick McDaniel, and Ian Goodfellow.
\newblock {Transferability in Machine Learning: from Phenomena to Black-Box
  Attacks using Adversarial Samples}.
\newblock {\em arXiv}, 2016.

\bibitem{10.1145/3052973.3053009}
Nicolas Papernot, Patrick McDaniel, Ian Goodfellow, Somesh Jha, Z~Berkay Celik,
  and Ananthram Swami.
\newblock {Practical Black-Box Attacks against Machine Learning}.
\newblock In {\em Proceedings of the 2017 ACM on Asia Conference on Computer
  and Communications Security}, ASIA CCS '17, page 506–519, New York, NY,
  USA, 2017. Association for Computing Machinery.

\bibitem{dr_ai_long}
Edward Raff, Shannon Lantzy, and Ezekiel~J Maier.
\newblock {Dr. AI, Where Did You Get Your Degree?}
\newblock In Fernando Koch, Andrew Koster, David Ria{\~{n}}o, Sara Montagna,
  Michael Schumacher, Annette ten Teije, Christian Guttmann, Manfred Reichert,
  Isabelle Bichindaritz, Pau Herrero, Richard Lenz, Beatriz L{\'{o}}pez, Cindy
  Marling, Clare Martin, Stefania Montani, and Nirmalie Wiratunga, editors,
  {\em Artificial Intelligence in Health}, pages 76--83, Cham, 2019. Springer
  International Publishing.

\bibitem{Raff2020a}
Edward Raff and Charles Nicholas.
\newblock {A Survey of Machine Learning Methods and Challenges for Windows
  Malware Classification}.
\newblock In {\em NeurIPS 2020 Workshop: ML Retrospectives, Surveys {\&}
  Meta-Analyses (ML-RSA)}, 2020.

\bibitem{Salman2020}
Hadi Salman, Andrew Ilyas, Logan Engstrom, Ashish Kapoor, and Aleksander Madry.
\newblock {Do Adversarially Robust ImageNet Models Transfer Better?}
\newblock {\em arXiv}, 2020.

\bibitem{Sebastian2016}
Marcos Sebasti{\'{a}}n, Richard Rivera, Platon Kotzias, and Juan Caballero.
\newblock {AVclass: A Tool for Massive Malware Labeling}.
\newblock In Fabian Monrose, Marc Dacier, Gregory Blanc, and Joaquin
  Garcia-Alfaro, editors, {\em Research in Attacks, Intrusions, and Defenses:
  19th International Symposium, RAID 2016}, pages 230--253. Springer
  International Publishing, Paris, France, 2016.

\bibitem{Sebastian2020}
Silvia Sebasti{\'{a}}n and Juan Caballero.
\newblock {AVClass2: Massive Malware Tag Extraction from AV Labels}.
\newblock In {\em ACSAC}, 2020.

\bibitem{Seideman2014}
Jeremy~D. Seideman, Bilal Khan, and Antonio~Cesar Vargas.
\newblock {Identifying malware genera using the Jensen-Shannon distance between
  system call traces}.
\newblock In {\em 2014 9th International Conference on Malicious and Unwanted
  Software: The Americas (MALWARE)}, pages 1--7. IEEE, 10 2014.

\bibitem{Shan2020Fawkes:Models}
Shawn Shan, Emily Wenger, Jiayun Zhang, Huiying Li, Haitao Zheng, and Ben~Y.
  Zhao.
\newblock {Fawkes: Protecting Privacy against Unauthorized Deep Learning
  Models}.
\newblock 2020.

\bibitem{Srivastava2014}
Nitish Srivastava, Geoffrey Hinton, Alex Krizhevsky, Ilya Sutskever, and Ruslan
  Salakhutdinov.
\newblock {Dropout : A Simple Way to Prevent Neural Networks from Overfitting}.
\newblock {\em The Journal of Machine Learning Research}, 15(1):1929--1958,
  2014.

\bibitem{10.1109/SP.2014.20}
Nedim Srndic and Pavel Laskov.
\newblock {Practical Evasion of a Learning-Based Classifier: A Case Study}.
\newblock In {\em Proceedings of the 2014 IEEE Symposium on Security and
  Privacy}, SP '14, page 197–211, USA, 2014. IEEE Computer Society.

\bibitem{217486}
Octavian Suciu, Radu Marginean, Yigitcan Kaya, Hal~Daume III, and Tudor
  Dumitras.
\newblock {When Does Machine Learning FAIL? Generalized Transferability for
  Evasion and Poisoning Attacks}.
\newblock In {\em 27th USENIX Security Symposium (USENIX Security 18)}, pages
  1299--1316, Baltimore, MD, 8 2018. {\{}USENIX{\}} Association.

\bibitem{Tramer2017}
Florian Tram{\`{e}}r, Nicolas Papernot, Ian Goodfellow, Dan Boneh, and Patrick
  McDaniel.
\newblock {The Space of Transferable Adversarial Examples}.
\newblock {\em arXiv}, pages 1--15, 2017.

\bibitem{Treede2018}
Rolf-Detlef Treede.
\newblock {The International Association for the Study of Pain definition of
  pain: as valid in 2018 as in 1979, but in need of regularly updated
  footnotes}.
\newblock {\em Pain reports}, 3(2):e643--e643, 3 2018.

\bibitem{Wong2020}
Eric Wong, Leslie Rice, and J.~Zico Kolter.
\newblock {Fast is better than free: Revisiting adversarial training}.
\newblock In {\em International Conference on Learning Representations (ICLR)},
  2020.

\bibitem{8953423}
C Xie, Z Zhang, Y Zhou, S Bai, J Wang, Z Ren, and A~L Yuille.
\newblock {Improving Transferability of Adversarial Examples With Input
  Diversity}.
\newblock In {\em 2019 IEEE/CVF Conference on Computer Vision and Pattern
  Recognition (CVPR)}, pages 2725--2734, 6 2019.

\bibitem{Zhou2018}
Wen Zhou, Xin Hou, Yongjun Chen, Mengyun Tang, Xiangqi Huang, Xiang Gan, and
  Yong Yang.
\newblock {Transferable Adversarial Perturbations}.
\newblock In Vittorio Ferrari, Martial Hebert, Cristian Sminchisescu, and Yair
  Weiss, editors, {\em ECCV}, pages 471--486, Cham, 2018. Springer
  International Publishing.

\end{thebibliography}
}

\appendix

\section{Figures referenced in main text}

\begin{figure}[!h]
\centering
\begin{subfigure}[b]{0.49\columnwidth}
\includegraphics[width=1\columnwidth]{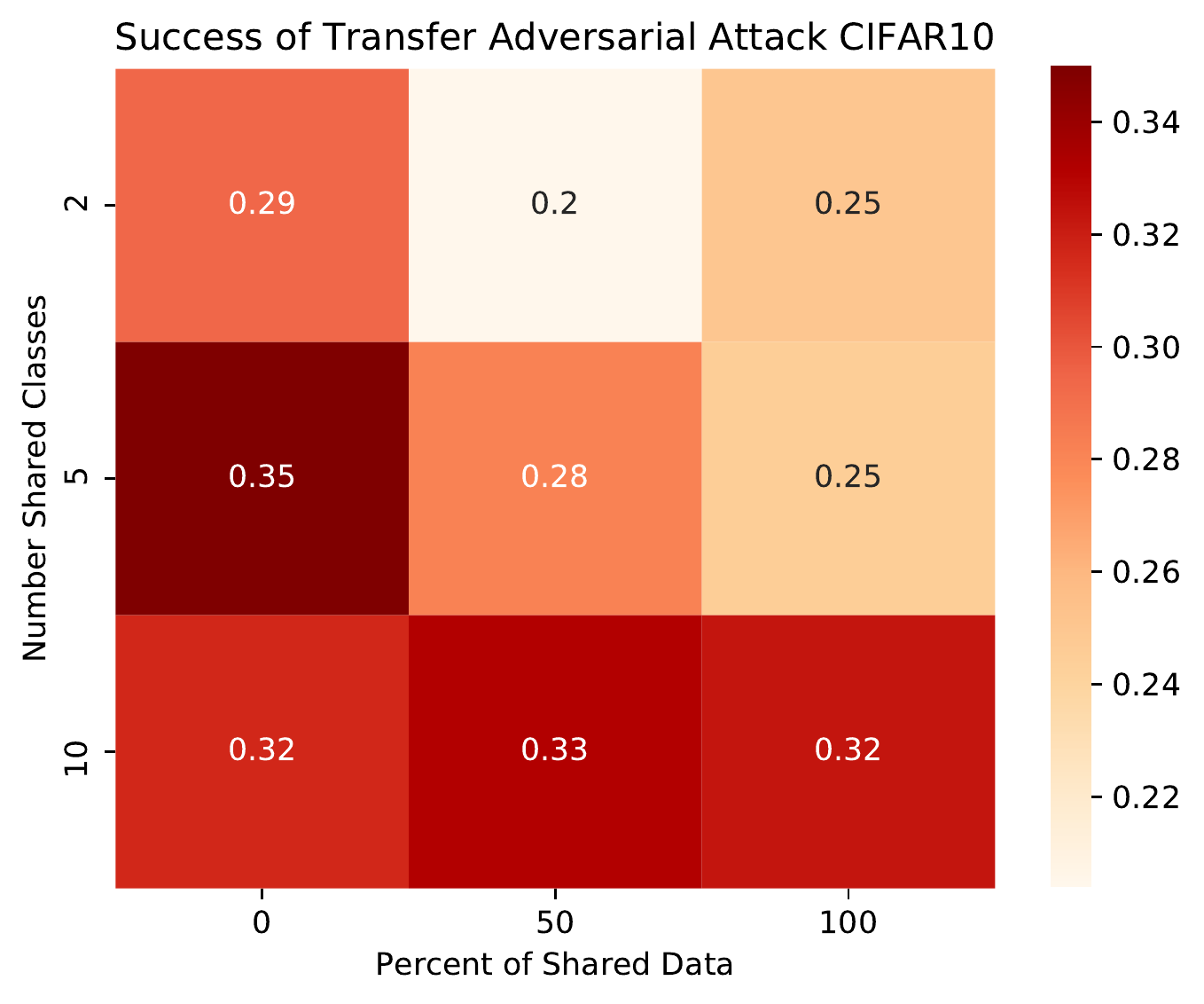}
\caption{
CIFAR-10}
\label{fig:CIFAR10CFNMTX}
\end{subfigure}
\begin{subfigure}[b]{0.49\columnwidth}
\centering
\includegraphics[width=1\columnwidth, scale=1.5]{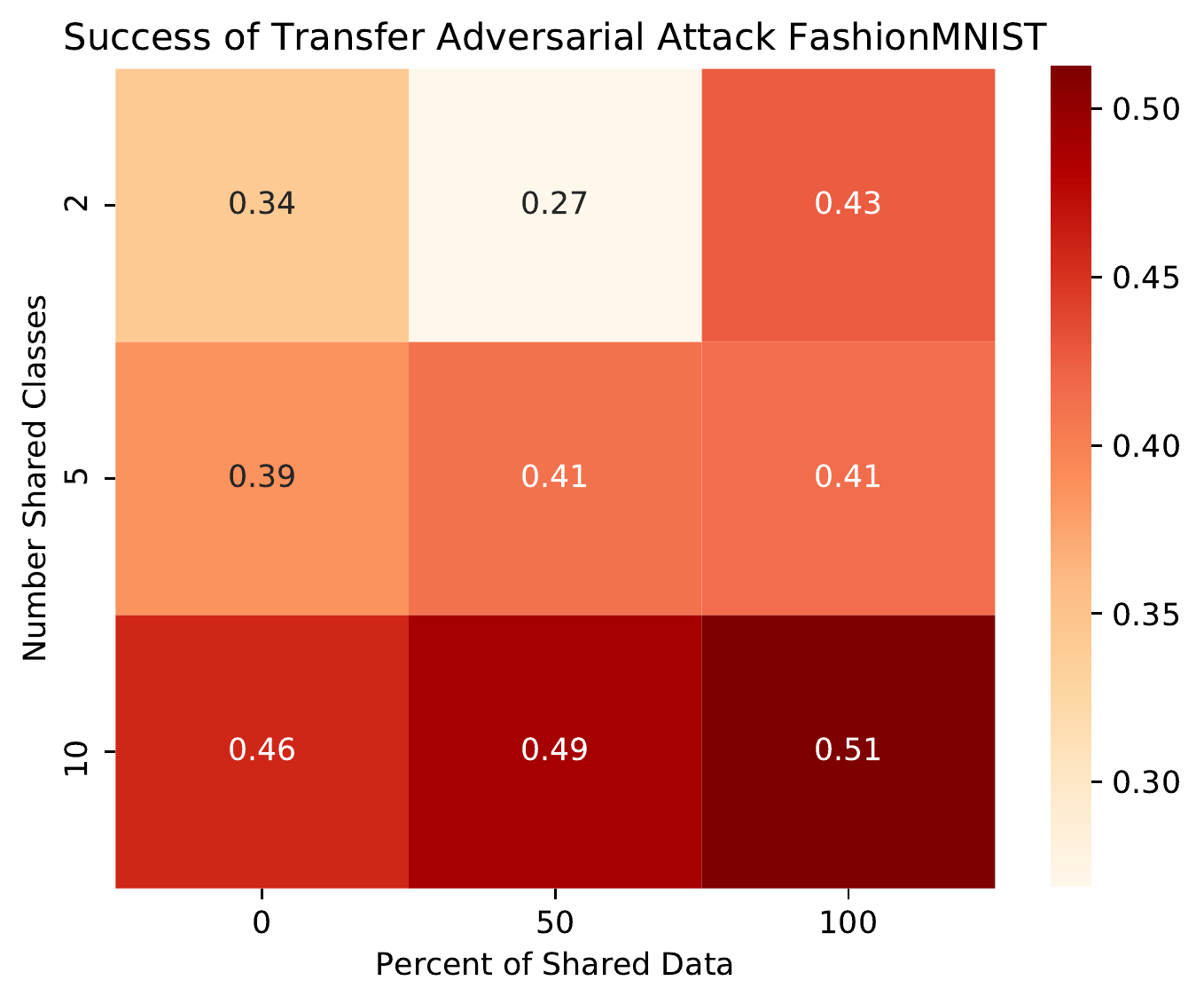}
\caption{
FashionMNIST}
\label{fig:FashionMNISTCFNMTX}
\end{subfigure}
\caption{Success of transfer attack dependent on variations of shared classes and percentage of data shared for two smaller datasets.}
\end{figure}

\end{document}